\documentclass[lettersize,journal]{IEEEtran}
\usepackage{amsmath,amsfonts}
\usepackage{algorithmic}
\usepackage{algorithm}
\usepackage{array}
\usepackage[caption=false,font=normalsize,labelfont=sf,textfont=sf]{subfig}
\usepackage{textcomp}
\usepackage{stfloats}
\usepackage{url}
\usepackage{verbatim}
\usepackage{graphicx}
\usepackage{cite}
\usepackage{tikz}
\usepackage{makecell}
\usepackage{booktabs}
\usepackage{amssymb}
\usepackage{xcolor}
\usetikzlibrary{positioning}
\usepackage{hyperref}
\begin{document}

\title{Multiclass Graph-Based Large Margin Classifiers: Unified Approach for Support Vectors \\and Neural Networks}

\author{
	Vítor~M.~Hanriot,
	Luiz~C.~B.~Torres,
	and~Antônio~P.~Braga
	\IEEEcompsocitemizethanks{
		\IEEEcompsocthanksitem Vítor~M.~Hanriot and Antônio~P.~Braga are with the Graduate Program in Electrical Engineering, Universidade Federal de Minas Gerais, Belo Horizonte 31270-901, Brazil (e-mail: vhanriot@ufmg.br; apbraga@ufmg.br).
		\IEEEcompsocthanksitem Luiz~C.~B.~Torres is with the Department of Computer and Systems, Universidade Federal de Ouro Preto, João Monlevade 35931-022, Brazil.
	}%
	\IEEEspecialpapernotice{\footnotesize In IEEE Transactions on Neural Networks and Learning Systems (2024).\\
	The final version is available at \href{https://doi.org/10.1109/TNNLS.2024.3420227}{https://doi.org/10.1109/TNNLS.2024.3420227}.}
}

\markboth{IEEE TRANSACTIONS ON NEURAL NETWORKS AND LEARNING SYSTEMS}%
{Shell \MakeLowercase{\textit{et al.}}: A Sample Article Using IEEEtran.cls for IEEE Journals}

\IEEEpubid{%
	\begin{minipage}{\textwidth}
		\vspace{1.5cm}        
		\centering
		\footnotesize
		© 2024 IEEE. Personal use of this material is permitted. Permission from IEEE must be obtained for all other uses, in any current or future media, including reprinting/republishing this material for advertising or promotional purposes, creating new collective works, for resale or redistribution to servers or lists, or reuse of any copyrighted component of this work in other works.
	\end{minipage}
}

\maketitle
\noindent

\begin{abstract}
While large margin classifiers are originally an outcome of an optimization framework, support vectors (SVs) can be obtained from geometric approaches. This article presents advances in the use of Gabriel graphs (GGs) in binary and multiclass classification problems. For Chipclass, a hyperparameter-less and optimization-less GG-based binary classifier, we discuss how activation functions and support edge (SE)-centered neurons affect the classification, proposing smoother functions and structural SV (SSV)-centered neurons to achieve margins with low probabilities and smoother classification contours. We extend the neural network architecture, which can be trained with backpropagation with a softmax function and a cross-entropy loss, or by solving a system of linear equations. A new subgraph-/distance-based membership function for graph regularization is also proposed, along with a new GG recomputation algorithm that is less computationally expensive than the standard approach. Experimental results with the Friedman test show that our method was better than previous GG-based classifiers and statistically equivalent to tree-based models.
\end{abstract}


\section{Introduction}
\IEEEPARstart{L}{arge} margin classifiers brought the perspective of classification learning being formalized not only as an empirical risk minimization problem but also as optimization of distances between the decision surface and margin vectors~\cite{vapnik1999nature}. Support Vector Machines (SVMs)~\cite{Vapnik_1992} yield the implementation of such an approach by adopting a quadratic programming (QP) formulation, which aims at maximizing the margin from Support Vectors (SVs), while considering the structural risk minimization principle or through solving a set of linear equations~\cite{suykens1999least,Carvalho_2009}.

With the recent success of deep neural networks (DNNs) on unstructured data~\cite{goodfellow2016deep} through the use of large datasets and extensive computing power~\cite{krizhevsky2017imagenet}, deep learning has been applied in the tabular data domain, including deep multilayer
perceptrons (MLPs)~\cite{kadra2021well}, ResNets~\cite{gorishniy2021revisiting} and Neural Oblivious Decision Ensembles~\cite{popov2019neural}. However, it has been reported that tree-based models still outperform DNNs for tabular data~\cite{grinsztajn2022tree,shwartz2022tabular}, specially XGBoost~\cite{chen2016xgboost}, LightGBM~\cite{ke2017lightgbm}, CatBoost~\cite{prokhorenkova2018catboost} and Random Forests~\cite{ho1995random}. SVMs remain present due to their performance on small to moderate-size datasets, which leads to constant improvements in the model, such as the proposal of new computation of slacks and kernels~\cite{vapnik2021reinforced} and twin SVMs~\cite{shao2023twin}, as well as new optimization formulations for multi-class problems such as adopting a linear programming approach~\cite{weston1998multi, carrasco2015multi}, penalty graphs~\cite{iosifidis2016multi}, and the decomposition algorithm method~\cite{gao2023multicycle}.

\begin{figure}[b]
\centering
\includegraphics[width=2.5in]{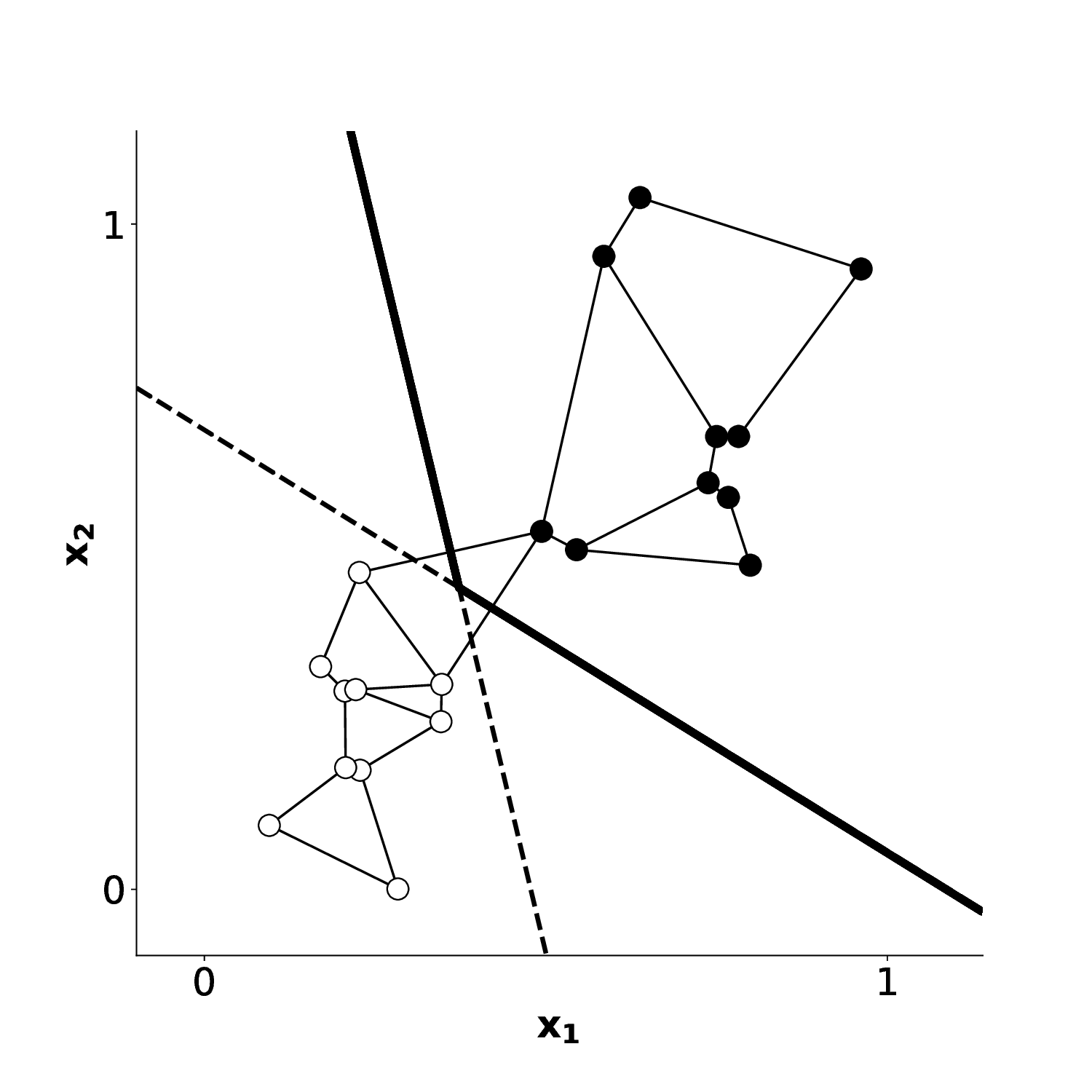}
\caption{Combination of 2 margin hyperplanes (dashed lines) resulting in Chipclass' decision boundary for a binary classification problem.}
\label{fig:chipclass_surface_with_planes}
\end{figure}

\IEEEpubidadjcol
Although SVs are originally an outcome of an optimization process, they can be obtained by considering the geometry of the data, once they can be seen as the points in the margin between the convex hulls of classes~\cite{bennett2000duality,peng2011geometric}. This geometric principle has been explored in previous works~\cite{Torres_2015, torres2022multi} in order to build classifiers that do not depend on explicit optimization nor on hyperparameters to be set in advance, which makes them suitable for autonomous learning and for applications that require less user-machine interaction, such as Internet of Things (IoT)~\cite{da2014internet} and edge computing~\cite{arias2022improved}. Chipclass~\cite{Torres_2015, arias2022improved} is based on the Gabriel Graph (GG)~\cite{Gabriel_1969}, an undirected graph that is constructed from Euclidean distance operations between pairs of the training set samples. From the extraction of the so-called Structural Support Vectors (SSVs), which are SVs obtained from the structural information of GG~\cite{zhang2002study}, Support Edges (SEs) are obtained, which are edges that connect SSVs from different classes, in order to define margin hyperplanes that are perpendicular to these edges. The final output of the classifier is obtained by a linear combination of such margin hyperplanes weighted by the distance between the test sample and the midpoint of the edge. Fig.~\ref{fig:chipclass_surface_with_planes} shows a dataset of a binary classification problem, the corresponding GG and the resulting margin hyperplanes that combined lead to the decision boundary, represented schematically in bold.

Another type of GG-based binary classifiers is based on the distance between the test sample and SSVs instead of SEs' midpoints: GG-based RBF networks (RBF-GG)~\cite{torres2014icann,queiroz2022rbf} have hidden layer neurons centered on SSVs, based on the principle of using SV-centered radial basis function~(RBF) Networks~\cite{scholkopf1997comparing}, and Gaussian Mixture Models (GMM-GG)~\cite{torres2020gmm} combine Gaussian distributions to yield final probability distributions for each class.

Both these approaches are studied in this paper, showing that SSV-centered classifiers present lower probabilities in the margin region and smoother classification contours. Moreover, it is shown in this work how smooth activation functions for Chipclass may prevent overlooking hidden layer neurons that are far from the test point due to the exponential decay factor of its original activation function.
Due to the studies on the architectures of these binary GG-based classifiers, a multi-class classifier based on SSV-centered hidden layer neurons with smooth activation functions is proposed.
At last, since the cardinality function used in Chiplcass, RBF-GG and GMM-GG is static and based solely on the adjacency submatrices defined by the neighborhood relationship of each sample, which makes different configurations with the same subgraph have the same membership function values, an extension to these filters is proposed using distance-based functions, which prove to be a generalization of the original filter. Since this extension requires a hyperparameter to define the radius of the distance-based kernels used for filtering, a new recomputation algorithm for GGs is presented.
Therefore, the contributions of this paper are:
\begin{itemize}
    \item Proposal of the use of smooth activation functions for GG-based classifiers.
    \item Study on previous GG-based architectures and empirical explanation on why SSV-centered activation functions yield better results.
    \item Extension of GG-based large margin classifiers to multi-class classification.
    \item Proposal of a distance-based membership function proven as a generalization of the cardinality function previously used in other GG-based classifiers, opening up new possibilities of filter policies.
    \item Recomputation of the GG in $\mathcal{O}(r(m-r)^2)$ instead of $\mathcal{O}((m-r)^3)$.
\end{itemize}
Experiments were carried out with 17 binary classification datasets from the UCI repository~\cite{lichman2013uci} and 15 multi-class classification from OpenML~\cite{vanschoren2014openml}. An ablation study was conducted to compare the proposed membership function and smooth activation functions for Chipclass, as well as a comparison between Chipclass, RBF-GG, GMM-GG and the proposed method. Moreover, we compare SSV-oriented Chipclass with k-Nearest Neighbors (kNN), SVMs, Random Forests, ResNets, XGBoost and LightGBM for binary and multi-class classification tasks. Experimental
results with the Friedman test showed that our method was statistically equivalent to models present in the literature.

The paper is organized as follows: in section~\ref{sec:background} we present Chipclass formulation; in Section~\ref{sec:motivation} we propose the methodology, which is divided into Chipclass improvement, SSV-oriented Chipclass proposition and multi-class SSV-oriented Chipclass. We describe how the experiments were done and show the results in Section~\ref{sec:results}. Our final considerations are presented in Section~\ref{sec:conclusion}.

\section{Background}
\label{sec:background}


\subsection{Gabriel Graph}
\label{subsec:gg_definition}
\begin{figure}[h]
\centering
\subfloat[]{\includegraphics[width=1.7in]{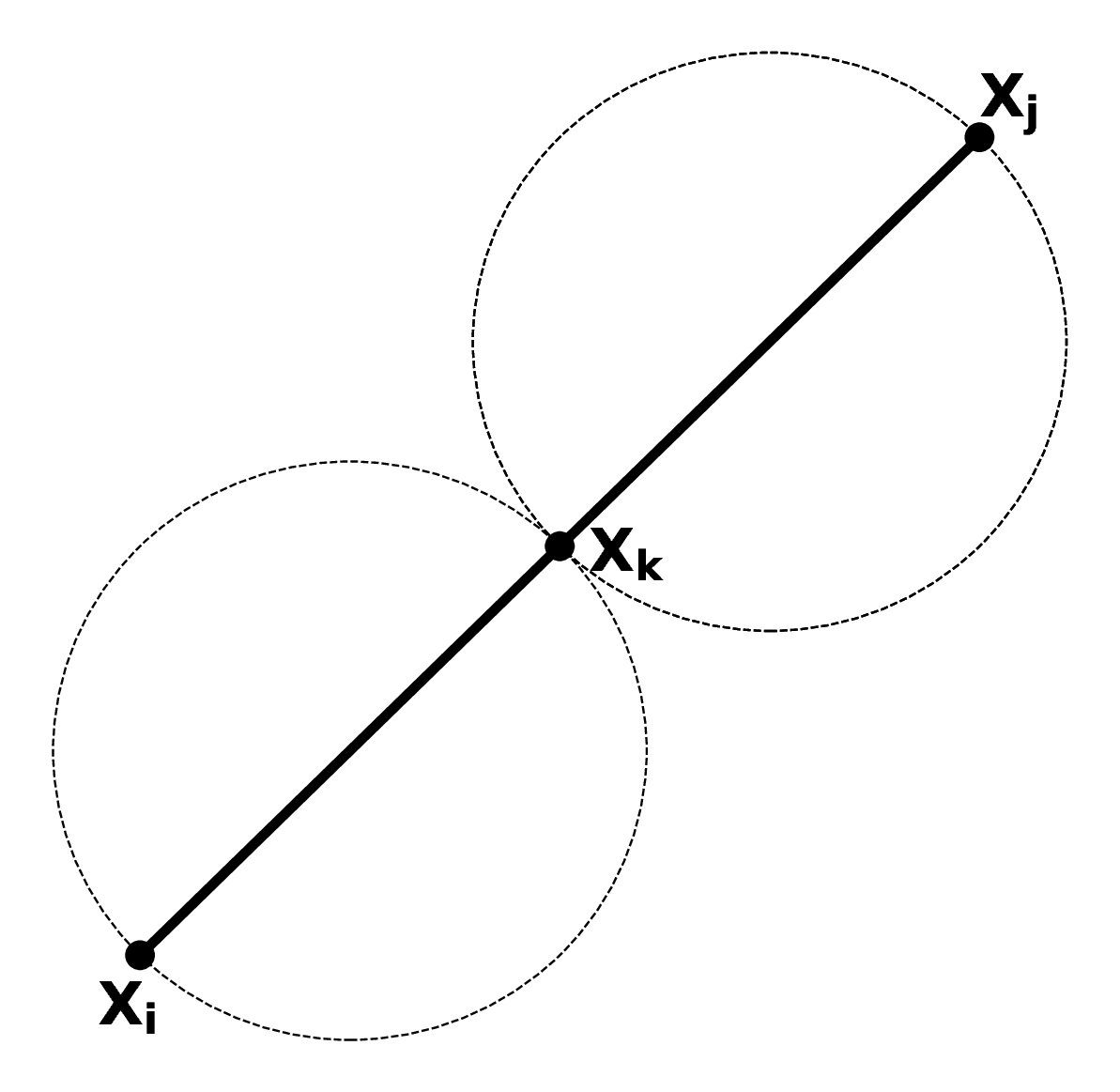}%
\label{fig:gg_xk_xj_edges}}
\hfil
\subfloat[]{\includegraphics[width=1.7in]{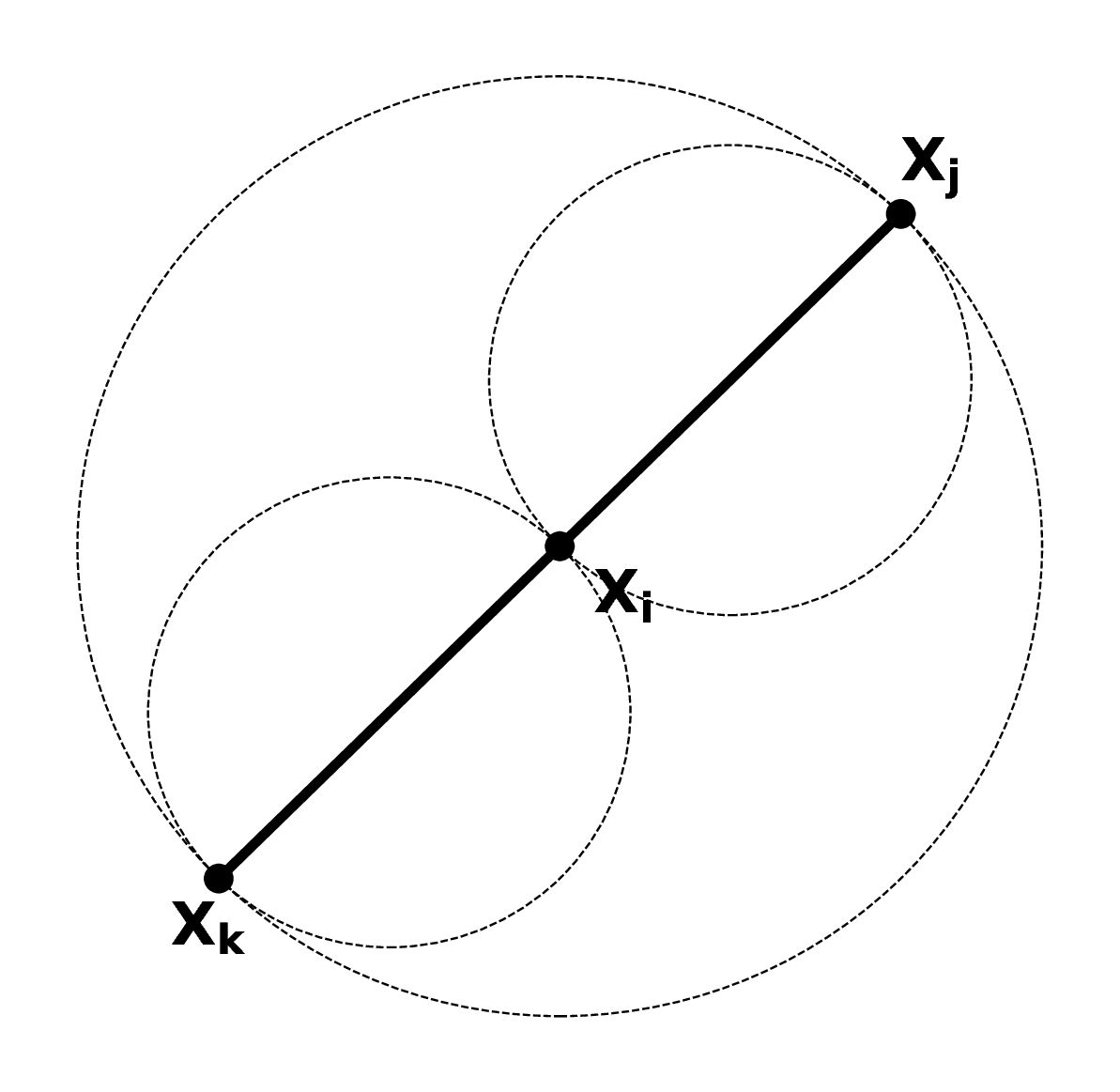}%
\label{fig:gg_xk_xj_notedges}}
\caption{($\textbf{X}_j$,$\textbf{X}_k$) $\in \mathcal{E}$ if Eq.~\ref{eq:gg_formulation} holds. (a) ($\textbf{X}_j$,$\textbf{X}_k$) $\in \mathcal{E}$ (b) ($\textbf{X}_j$,$\textbf{X}_k$) $\notin \mathcal{E}$.}
\label{fig:avg_margin_threshold}
\end{figure}

\begin{figure}[h]
\centering
\includegraphics[width=3in]{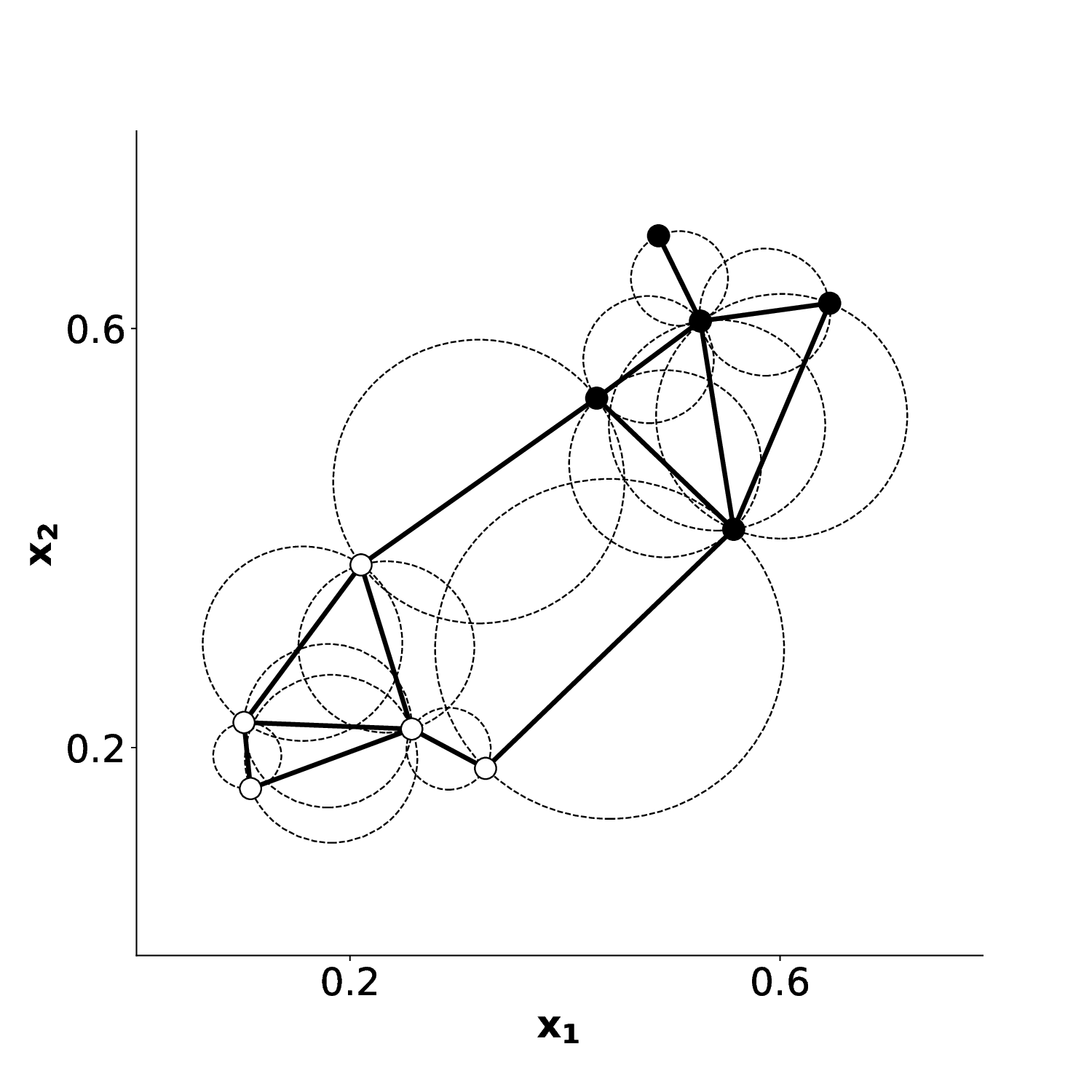}
\caption{Gabriel Graph for a binary classification problem, all 2-spheres that follow Eq.~\ref{eq:gg_formulation} are illustrated.}
\label{fig:gg_formulation}
\end{figure}

Given a set of finite samples $\mathcal{S} = \{(\textbf{X}_i,\textbf{y}_i) \in \mathcal{X} \times \mathcal{Y}\}_{i=1}^{m}$, the Gabriel Graph (GG) of $\mathcal{S}$ is an undirected graph in which a pair of vertices $\textbf{X}_j$ and $\textbf{X}_k$ from $\mathcal{V} = \{\textbf{X}_i\}_{i=1}^m$ define an edge ($\textbf{X}_j$,$\textbf{X}_k$) $\in \mathcal{E}$ if the condition of Eq.~\ref{eq:gg_formulation} is met.
\begin{equation}
\begin{split}
    ||\textbf{X}_j-\textbf{X}_k||^2 & \le (||\textbf{X}_j-\textbf{X}_i||^2+||\textbf{X}_k-\textbf{X}_i||^2) \\ 
    & \;\;\;\;\;\;\; \forall \; i=1,...,m \; | \; i \neq j \neq k \neq i
\end{split}
    \label{eq:gg_formulation}
\end{equation}

\noindent where $||\cdot||$ is the Euclidean distance between two samples~\cite{Gabriel_1969}. Thus, $\textbf{X}_j$ and $\textbf{X}_k$ are connected in the graph only if no other sample from $\mathcal{S}$ is within the $D$-sphere with center $\frac{\textbf{X}_j+\textbf{X}_k}{2}$ and diameter $||\textbf{X}_j-\textbf{X}_k||$. Figs.~\ref{fig:gg_xk_xj_edges} and~\ref{fig:gg_xk_xj_notedges} show examples of whether or not ($\textbf{X}_j$,$\textbf{X}_k$) define an edge, given a third sample $\textbf{X}_i$. Fig.~\ref{fig:gg_formulation} depicts a GG and all 2-spheres that follow Eq.~\ref{eq:gg_formulation}.
\subsection{GG's structural information}
\label{subsec:gg_ses_ssvs}
\begin{figure}[h]
\centering
\includegraphics[width=3in]{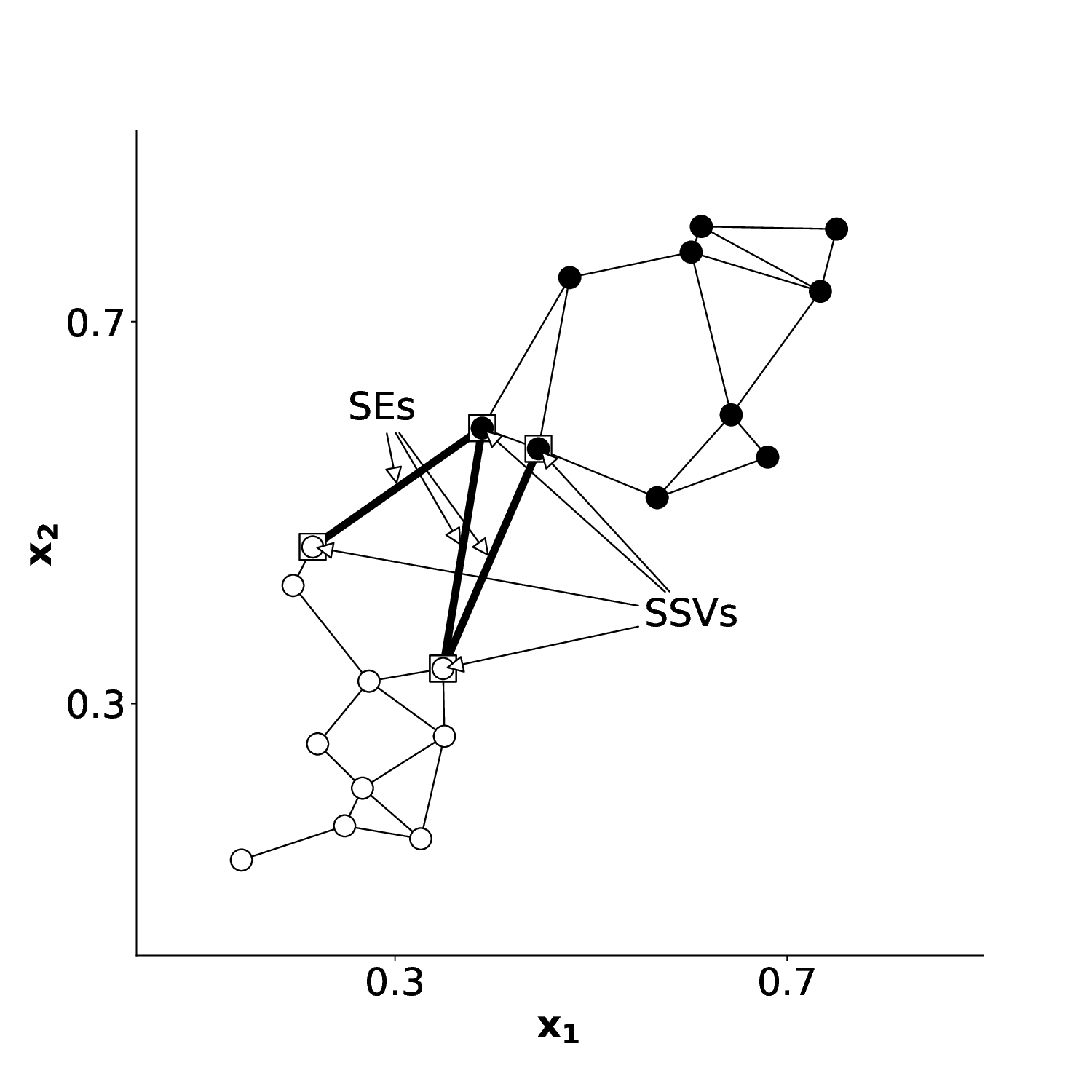}
\caption{SSVs and SEs highlighted for a GG obtained from a set of samples of a binary classification problem.}
\label{fig:gg_ssvs_ses}
\end{figure}

Once GG's vertices $\mathcal{V}$ and edges $\mathcal{E}$ are computed, the vertices $(\textbf{X}_j, \textbf{X}_k) \in \mathcal{E}$  are called Structural Support Vectors (SSVs) if $\textbf{y}_k \neq \textbf{y}_j$\cite{zhang2002study}. If so, then $(\textbf{X}_j, \textbf{X}_k)$ is called a Support Edge (SE)~\cite{Torres_2015}. Fig.~\ref{fig:gg_ssvs_ses} highlights the SSVs and SEs of a binary classification problem.
\subsection{Chipclass}
\label{subsec:chipclass_definition}
Let $(\textbf{X}_j, \textbf{X}_k)$ be an SE, then the hyperplane that contains $\frac{(\textbf{X}_j+ \textbf{X}_k)}{2}$ with normal vector $\frac{\textbf{X}_j- \textbf{X}_k}{||\textbf{X}_j-\textbf{X}_k||}$ defines a maximum margin classifier between SSVs $\textbf{X}_j$ and $\textbf{X}_k$. Chipclass~\cite{Torres_2015} is then defined as a single hidden layer neural network, composed by such hyperplanes, having the expression represented in Eq.~\ref{eq:gating} as the output activation function.
\begin{equation}
    \text{h}_k(\textbf{x}) = \text{exp}\left(\frac{\text{max}(||\textbf{x}-\textbf{P}_i||)^2}{||\textbf{x}-\textbf{P}_k||}\right) \; \forall \; i= 1,...,m
    \label{eq:gating}
\end{equation}

\noindent where $m$ is the number of hyperplanes defined by the SEs of the training set, $\textbf{P}_k$ is the middle point between the SSVs that define the $k$th hyperplane and all $\text{h}_k$ are normalized afterwards so that $\sum\limits_{k=1}^m\text{h}_k(\textbf{x})=1$. 

Therefore, the closer the hyperplane to the test sample, the greater its contribution to the final classification. The class that will benefit most from such distance contribution, assigned to the weight $\textbf{w}_k$ of the $k$th hyperplane, is the class of the SSV closest to the test sample. The weight assignment is represented in Eq.~\ref{eq:definition_cpcn}, where $\alpha_k$ and $\beta_k$ are, respectively, the positive and negative SSVs from the $k$th hyperplane.
\begin{equation}
  \textbf{w}_k =\begin{cases}
  1,& \text{if } ||\textbf{x}-\alpha_k|| < ||\textbf{x}-\beta_k||\\
  -1,              & \text{otherwise.}
\end{cases}
\label{eq:definition_cpcn}
\end{equation}

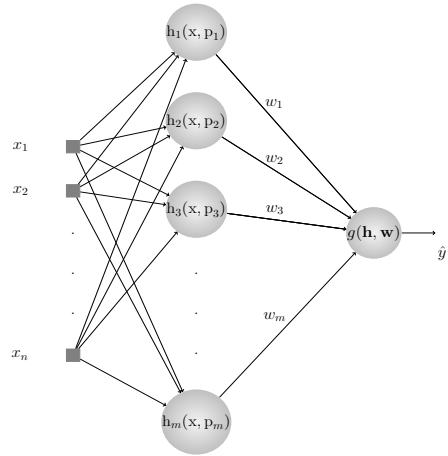
\begin{figure}[h]
	\centering
	\resizebox{6cm}{6cm}{
		\begin{tikzpicture}
			[rectin/.style={rectangle,draw=gray,fill=gray,thick,
				inner sep=0pt,minimum size=4mm},
			rectout/.style={rectangle,draw=none,fill=none,
				inner sep=0pt,minimum size=4mm},
			neuro/.style={circle,shade,inner color=gray!10,outer color=gray!50,thin,
				inner sep=0pt,minimum size=8mm},
			texto/.style={}]
			
			\node[rectin] (x1)       {};
			\node[rectin] (x2)  [below=of x1]     {};
			\node[] (x1txt)  [left=of x1]     {\Large $x_1$};
			\node[] (x2txt)  [left=of x2]     {\Large $x_2$};
			\node[] (pi1)  [below=of x2]     {\Large{.}};
			\node[] (pi2)  [below=of pi1]     {\Large{.}};
			\node[] (pi3)  [below=of pi2]     {\Large{.}};
			\node[rectin] (xn)  [below=of pi3]     {};
			\node[] (xntxt)  [left=of xn]     {\Large $x_n$};
			\node[neuro] (n1)  [above right=4cm of x1]     {\Large $\text{h}_1(\textbf{x}, \textbf{p}_1)$};
			\node[neuro] (n2)  [below= of n1]     {\Large $\text{h}_2(\textbf{x}, \textbf{p}_2)$};
			\node[neuro] (n3)  [below= of n2]     {\Large $\text{h}_3(\textbf{x}, \textbf{p}_3)$};
			
			\node[] (p1)  [below=of n3]     {\Large{.}};
			\node[] (p2)  [below=of p1]     {\Large{.}};
			\node[] (p3)  [below=of p2]     {\Large{.}};
			
			\node[neuro] (n5)  [below= of p3]     {\Large $\text{h}_m(\textbf{x}, \textbf{p}_m)$};

			\draw[->] (x1) edge (n1);
			\draw[->] (x1) edge (n2);
			\draw[->] (x1) edge (n3);
			
			\draw[->] (x1) edge (n5);

			\draw[->] (x2) edge (n5);

			\draw[->] (xn) edge (n5);

			\node[neuro] (nsaida)  [right= 8cm of pi1]     {\Large $g(\mathbf{h},\mathbf{w})$};

			\draw[->] (n1) edge (nsaida);
			\draw[->] (n2) edge (nsaida);
			\draw[->] (n3) edge (nsaida);
			
 			\draw[->] (n5) edge (nsaida);

			\node[rectout] (saida)  [right= of nsaida]     {};

			\draw[->] (x2) -- (n1);
			\draw[->] (x2) -- (n2) ;
			\draw[->] (x2) -- (n3);

			\draw[->] (xn) -- (n1);
			\draw[->] (xn) -- (n2) ;
			\draw[->] (xn) -- (n3) ;

			\draw[thick] (n1) -- (nsaida) node[midway,above = 0.7cm] {\Large $w_1\;\;\;\;\;$};
			\draw[thick] (n2) -- (nsaida) node[midway,above=0.3cm] {\Large $w_2\;\;\;\;\;$};
			\draw[thick] (n3) -- (nsaida) node[midway,above=0.1cm] {\Large $w_3\;\;\;\;\;$};
			\draw[thick] (n3) -- (nsaida) node[midway,below=2.8cm] {\Large $w_m\;\;\;\;\;$};

			\draw[->] (nsaida) edge (saida);
			\node[] (fx)  [below=0.1cm of saida]     {\Large $\hat{y}$};

		\end{tikzpicture}

	}
	\caption{Schematic representation of Chipclass: a single hidden layer neural network.}
	\label{fig:chipclass_onehiddenlayer}
\end{figure}

Thus, the classifier can be seen as a single hidden layer neural network (Fig.~\ref{fig:chipclass_onehiddenlayer}), in which each hidden layer neuron corresponds to a hyperplane and the output layer is a linear combination of these maximum margin classifiers, with a sigmoidal activation function at the output, so that its response is the probability of one of the classes, given the aggregation of all contributions, as shown in Eq.~\ref{eq:prob_chipclass}. 
\begin{equation}
    \mathbb{P}(y=1|\textbf{x}) = \text{sigmoid}(\sum\limits_{k=1}^h \textbf{w}_k \cdot \text{h}_k(\textbf{x}))
\label{eq:prob_chipclass}
\end{equation}
    
\subsubsection{Chipclass' Regularization}
In order to reduce the effects of overfitting due to class overlapping, such GG-based classifiers rely on their own graph to assess topological quality of the data and labeling coherence of the training set~\cite{aupetit2005high}. Such information is considered for reducing the output response of the model.

Given the dataset $\mathcal{S}$ and its associated GG $G(\mathcal{S})$, the membership (quality) value of a sample $(\textbf{X}_i,\textbf{y}_i) \in \mathcal{S}$ is defined in Eq.~\ref{eq:membership_value}.
\begin{equation}
    \text{q}(\textbf{X}_i) = \frac{|{(\textbf{X}_k,\textbf{y}_k) \in G(\mathcal{S})(\textbf{X}_i) \; | \; \textbf{y}_k=\textbf{y}_i }| }{|(\textbf{X}_k,\textbf{y}_k) \in G(\mathcal{S})(\textbf{X}_i)|}
    \label{eq:membership_value}
\end{equation}

\noindent where $|\cdot|$ is the cardinality of a set and $G(\mathcal{S})(\textbf{X}_i)$ the subgraph of GG that only contains the neighbors of $\textbf{X}_i$, so Eq.~\ref{eq:membership_value} is the ratio between the number of neighbors of $\textbf{X}_i$ labeled as $\textbf{y}_i$ and the total number of neighbors of $\textbf{X}_i$.

Thus, the examples of class $c$ with membership value lower than the threshold $t_c$ (Eq.~\ref{eq:membership_threshold}), which was originally defined~\cite{Torres_2015} as the mean of the membership values for all the samples that are labeled as $c$ ($\mathcal{Q}_c$), are filtered from the original GG. Then, a new GG is computed after filtering, SEs and SSVs are found and then the classifier's architecture presented in subsec.~\ref{subsec:chipclass_definition} is defined. The effect of such sample removal is the elimination of those hyperplanes that would cause higher complexity terms, and thus overfitting, in the final function.
\begin{equation}
    t_c = \frac{\sum_{q_i \in \mathcal{Q}_c} q_i}{|\mathcal{Q}_c|}
    \label{eq:membership_threshold}
\end{equation}

\section{Methodology}
\label{sec:motivation}
For GG-distance-based classifiers, new activation functions and a new distance-based filter for regularization are proposed in subsec.~\ref{subsec:improving_chipclass}. It is discussed the effect of Chipclass near the margin and presented an SSV-oriented neural network architecture for Chipclass (subsec.~\ref{subsec:architecture_svvoriented}). The proposed architecture is then compared with GG-based RBF networks (RBF-GG)~\cite{torres2014icann} and Gaussian Mixture Models (GMM-GG)~\cite{torres2020gmm} classifiers that also rely on SSVs but need GG's structural information to define the hyperparameters of such models. The principles of the new architecture are then extended for multi-class classification problems in subsec.~\ref{subsec:multiclass_classification}.

\subsection{Improving Chipclass}
\label{subsec:improving_chipclass}
\subsubsection{Distance-based activation function}


Considering Chipclass' activation function given in Eq.~\ref{eq:gating}, let
$m_d = \text{max}(||\textbf{x}-\textbf{P}_i||)$  ($\forall \; i= 1,...,m$), its derivative is presented in Eq.~\ref{eq:derivative_actfunchip}.
\begin{equation}
	\text{h}_k'(\textbf{x}) = \frac{\partial \text{h}_k(\textbf{x})}{\partial (||\textbf{x}-\textbf{P}_k||)} = -\frac{m_d^2~\text{exp}\left(\frac{m_d^2}{||\textbf{x}-\textbf{P}_k||}\right)}{(||\textbf{x}-\textbf{P}_k||)^2}
	\label{eq:derivative_actfunchip}
\end{equation}

\begin{figure}[h]
\centering
\includegraphics[width=2in]{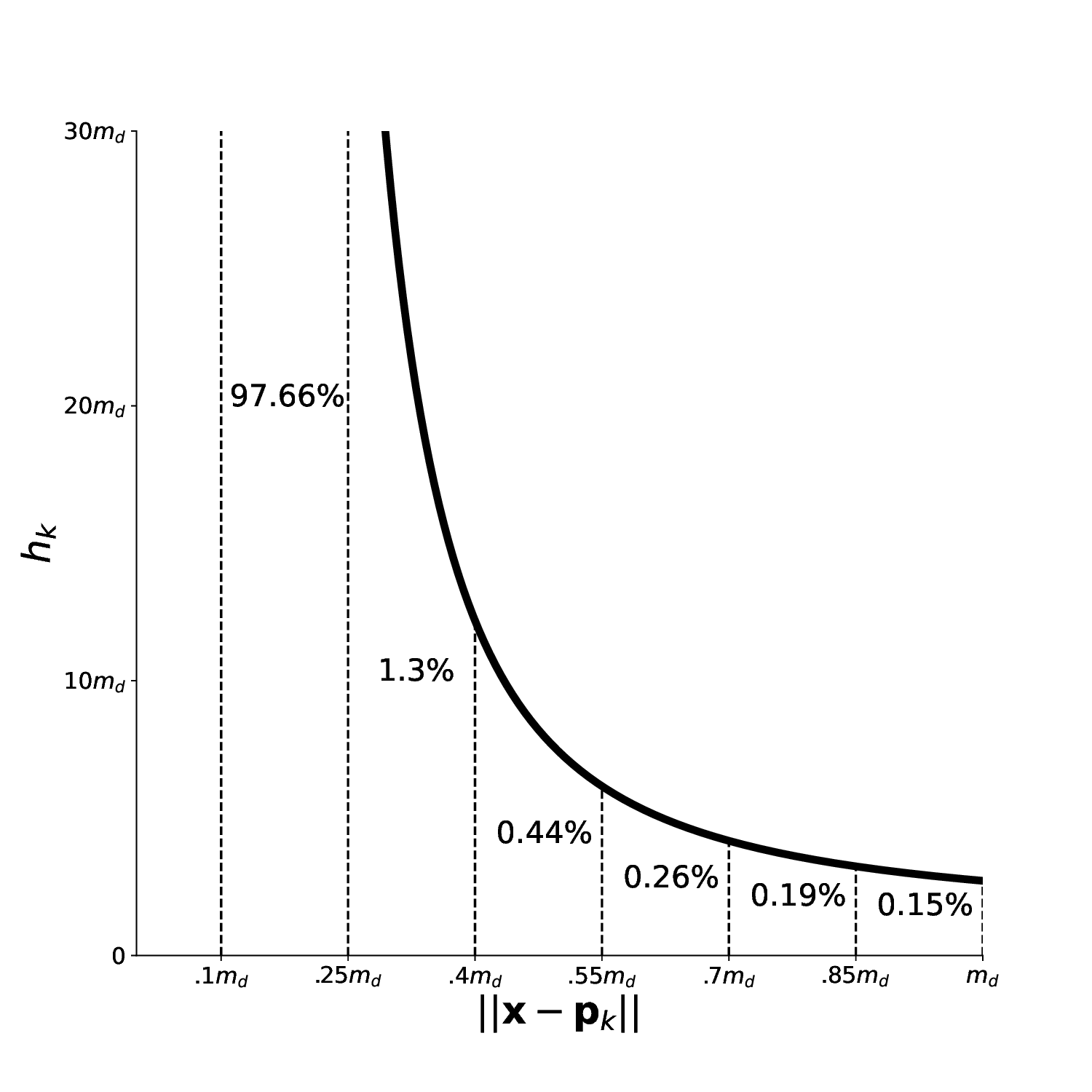}
\caption{Chipclass activation function with respect to $m_d$: areas under the curve for $.15m_d$ intervals in the range of [$0.1m_d$, $m_d$] are highlighted.}
\label{fig:chipclass_actfun}
\end{figure}

As it can be observed in Fig.~\ref{fig:chipclass_actfun}, which presents Chipclass' activation function with respect to $m_d$, $\text{h}_k$ is highly sensitive with respect to $||\textbf{x}-\textbf{P}_k||$, so that $\text{h}_k' \to -\infty$ when $ ||\textbf{x}-\textbf{P}_k|| \to 0^+$. Likewise, 
\begin{equation*}
    \lim_{||\textbf{x}-\textbf{P}_k||\to0^+} \text{h}_k(\textbf{x}) = \infty
\end{equation*}

\noindent so that the ratio between the areas under the curve for the intervals [0, $0.1m_d$] and [$0.1m_d$, $m_d$] can be described as in Eq.~\ref{eq:ratio_integrals}.
\begin{equation}
    \frac{\int_{0}^{0.1m_d} \text{h}_k(\textbf{x}) \,\partial(||\textbf{x}-\textbf{P}_k||)}{\int_{0.1m_d}^{m_d} \text{h}_k(\textbf{x}) \,\partial(||\textbf{x}-\textbf{P}_k||)} \to \infty
    \label{eq:ratio_integrals}
\end{equation}
 
Thereby, the percentages of the area under the curve for $0.15m_d$ intervals in the range of [$0.1m_d$, $m_d$] are also presented in Fig.~\ref{fig:chipclass_actfun}.

Unlike radial-basis Gaussian functions or other common activation functions used in ML such as $\text{tanh}$ and $\text{sigmoid}$, $\text{h}_k(\textbf{x})$'s density is highly concentrated for values of $||\textbf{x}-\textbf{P}_k||$ closer to 0. Therefore, given two middle points $\textbf{P}_j$ and $\textbf{P}_k$ that define two hidden layer neurons, if $||\textbf{x}, \textbf{P}_j|| > ||\textbf{x}-\textbf{P}_k||$, then $h_j \ll \text{h}_k(\textbf{x})$ the closer $\textbf{P}_k$ is to $\textbf{x}$. This may lead to a frequent disregard of hidden layer neurons which have their centers further away from $\textbf{x}$.

Thus, it is proposed in this paper the use of smoother activation functions, such as the one presented in Eq~\ref{eq:gating_tanh}. For $\text{h}_{ktanh}(\textbf{x})$, an offset of $+1$ is added so that $0 < \text{h}_{ktanh}(\textbf{x}) \leq 1$. After applying normalization, $\sum\limits_{k=1}^m\text{h}_{ktanh}(\textbf{x})=1$.
\begin{equation}
    \text{h}_{ktanh}(\textbf{x}) = \text{tanh}(-||\textbf{x}-\textbf{P}_k||)+1
    \label{eq:gating_tanh}
\end{equation}

\subsubsection{Distance-Based Filter}
As discussed in Section~\ref{subsec:chipclass_definition}, regularization in Chipclass is accomplished by removing uncertain samples according to class representation in their neighborhood sub-matrix. The original approach, however, considers only adjacencies in the graph, thus different arrangements of data that generate the same sub-matrix of adjacencies may lead to the same value of quality $\text{q}(\textbf{X}_i)$. Therefore, it is considered in this paper the weighting of the membership function based on the distances of each graph neighbor to $\textbf{X}_i$ (Eq.~\ref{eq:membership_value_dist}).
\begin{equation}
    \text{q}_d(\textbf{X}_i) = \frac{\sum_{\textbf{y}_k = \textbf{y}_i}\mathcal{K}(\textbf{X}_i, \textbf{X}_k)}{\sum\mathcal{K}(\textbf{X}_i, \textbf{X}_k)}  \; \forall \; \textbf{X}_k \in G(\mathcal{S})(\textbf{X}_i)
    \label{eq:membership_value_dist}
\end{equation}

\noindent where $\mathcal{K}$ is the Gaussian kernel defined in Eq.~\ref{eq:gaussian_kernel},
\begin{equation}
    \mathcal{K}(\textbf{X}_i, \textbf{X}_k) = \text{exp}\left(-\frac{||\textbf{X}_i-\textbf{X}_k||^2}{2\sigma^2}\right)
    \label{eq:gaussian_kernel}
\end{equation}

\noindent where $\sigma$ is predefined. 

\begin{figure}[h]
\centering
\subfloat[]{\includegraphics[width=1.15in]{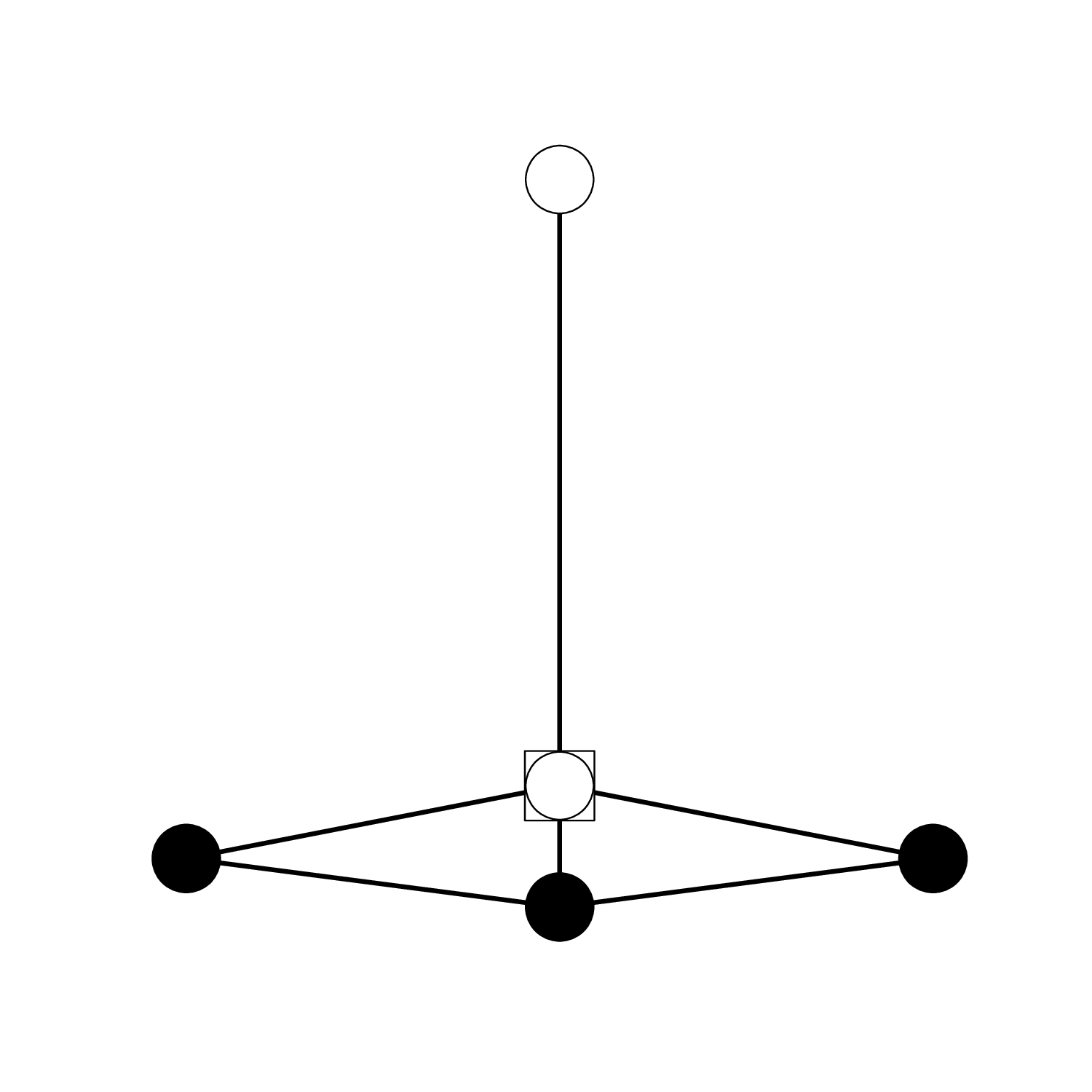}%
\label{fig:qdless}}
\hfil
\subfloat[]{\includegraphics[width=1.15in]{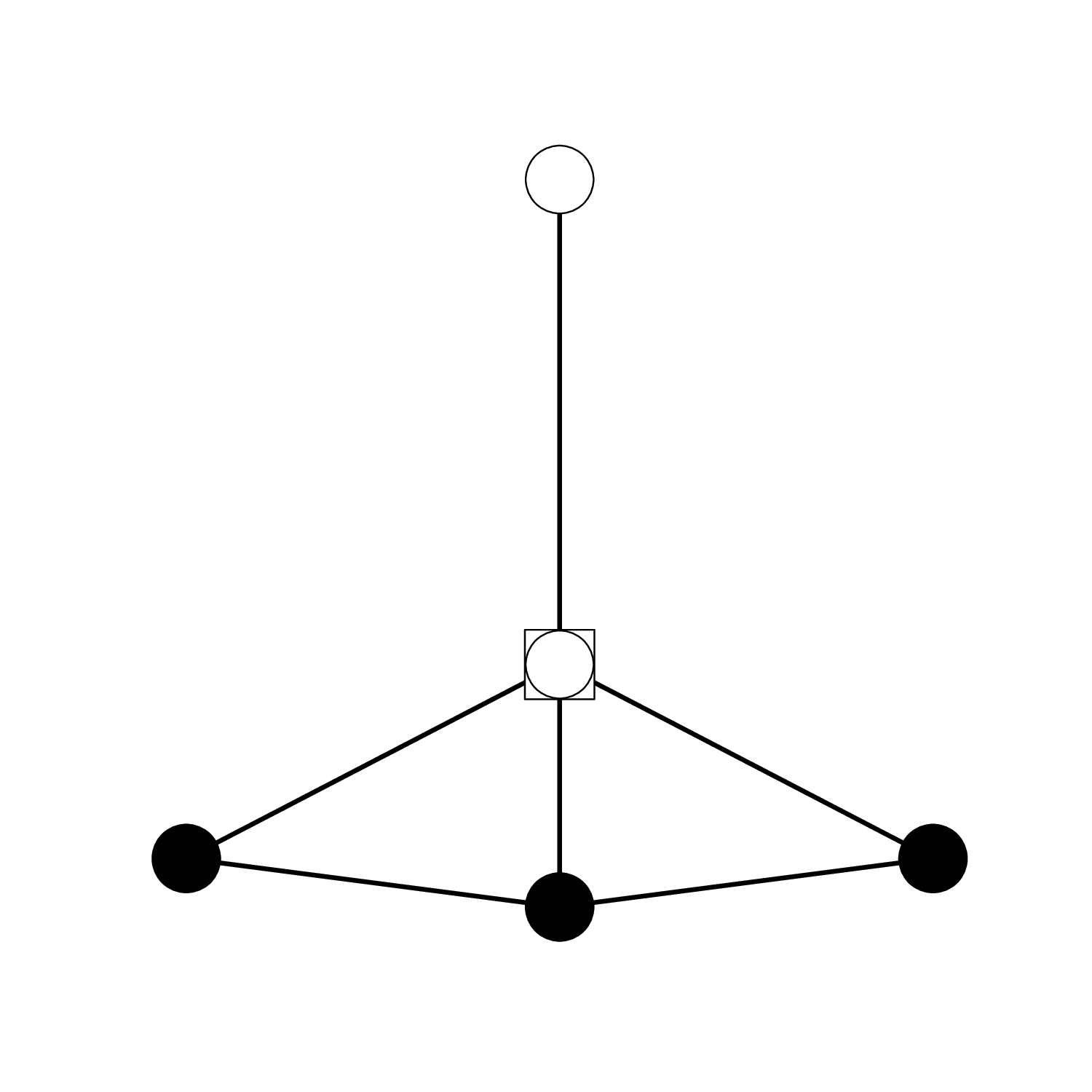}%
\label{fig:qdapprox}}
\hfil
\subfloat[]{\includegraphics[width=1.15in]{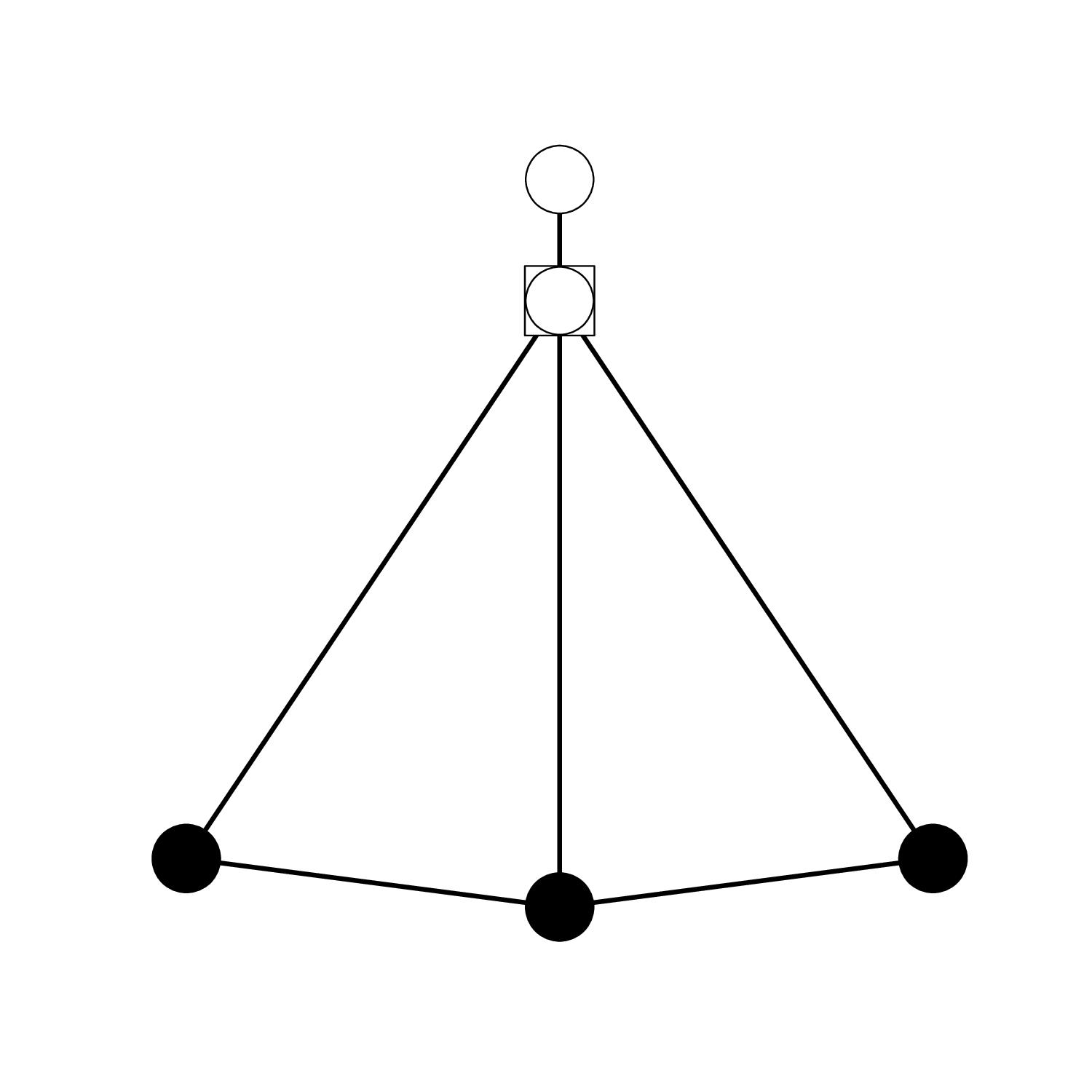}%
\label{fig:qdlarger}}
\caption{Subgraph $G(\mathcal{S})(\textbf{X}_i)$ is the same for different arrangements of $\textbf{X}_i$, therefore $\text{q}(\textbf{X}_i)$ holds the same value, on the contrary $\text{q}_d(\textbf{X}_i)$ varies with the distances. (a) $\text{q}_d(\textbf{X}_i) < q(\textbf{X}_i)$. (b) $\text{q}_d(\textbf{X}_i) \approx q(\textbf{X}_i)$. (c) $\text{q}_d(\textbf{X}_i) > q(\textbf{X}_i)$.}
\end{figure}

As an example, Figs.~\ref{fig:qdless} to~\ref{fig:qdlarger} illustrate the subgraphs $G(\mathcal{S})(\textbf{X}_i)$ for different arrangements of $\textbf{X}_i$, which is highlighted with a square in the figures. While for all configurations $\text{q}(\textbf{X}_i) = \frac{1}{4}$ since the adjacency matrix remains the same for all cases, the new quality index $\text{q}_d(\textbf{X}_i)$ for a fixed low $\sigma$ value increases as $\textbf{X}_i$ gets closer to its neighbor of the same class and moves away from the neighbors of different classes. The consideration of distances in addition to the neighborhood sub-matrix may cope with situations like the ones in the figures, which particularly appear with sparse datasets.

\begin{figure}[h]
\centering
\subfloat[]{\includegraphics[width=1.73in]{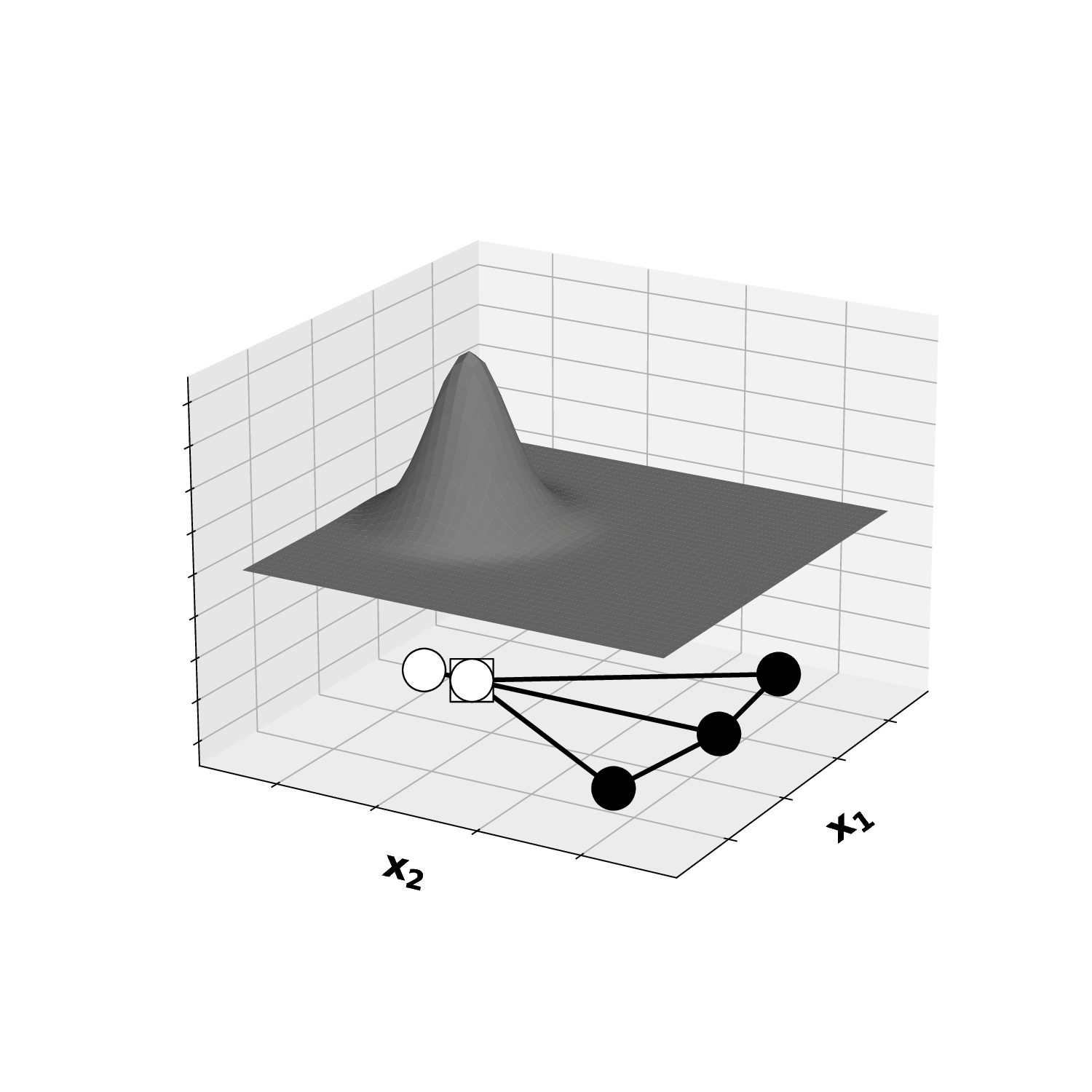}%
\label{fig:sigma_comp_low}}
\hfil
\subfloat[]{\includegraphics[width=1.73in]{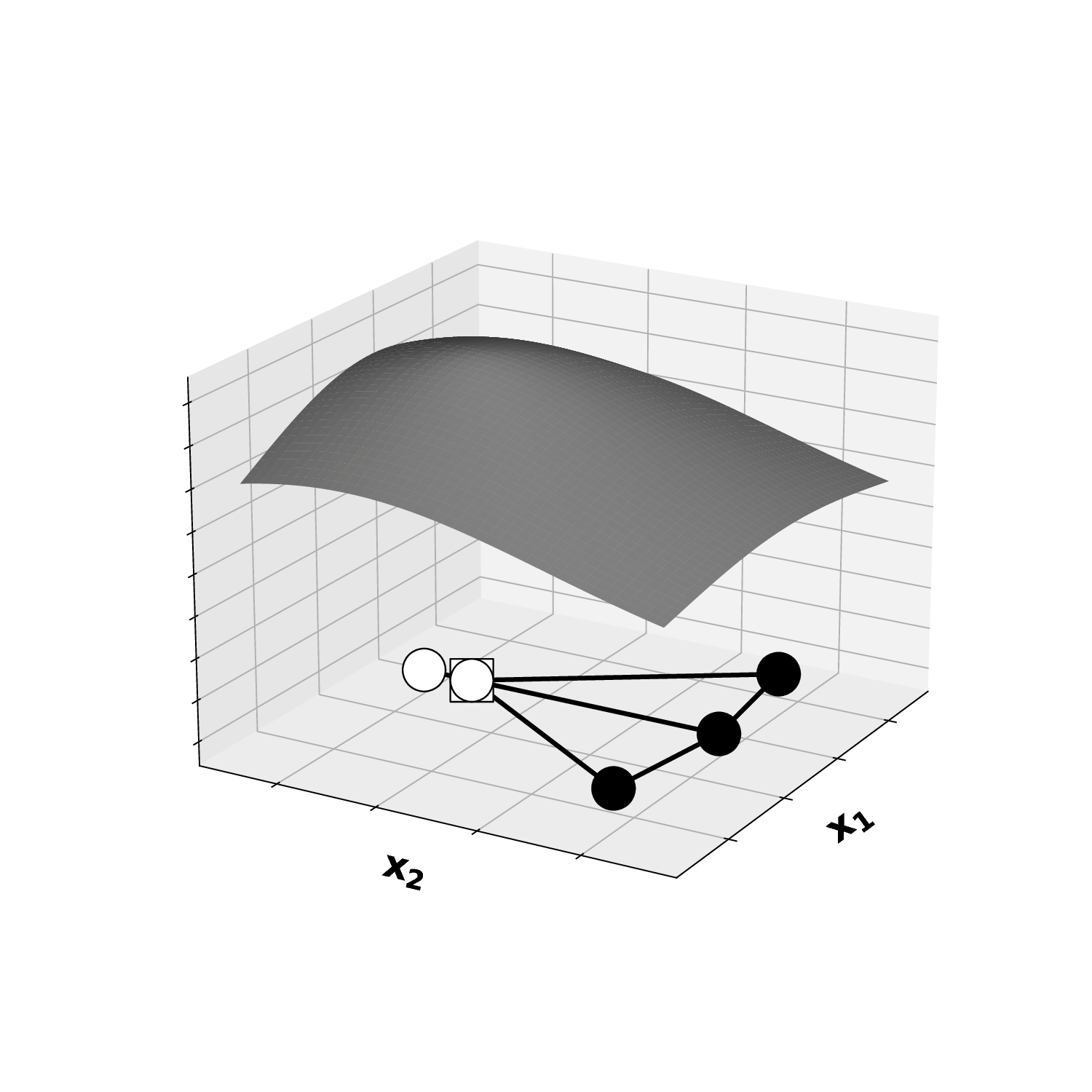}%
\label{fig:sigma_comp_high}}
\caption{Kernel values of Eq.~\ref{eq:gaussian_kernel} for different $\sigma$ values for the highlighted sample in square. (a) Low $\sigma$ prioritizes the sample of the same class, thus $\text{q}_d(\textbf{X}_i)$ is higher. (b) High $\sigma$ makes distance to the neighbors less important, thus $\text{q}_d(\textbf{X}_i)$ is lower.}
\end{figure}

Besides that, Eq.~\ref{eq:gaussian_kernel}'s $\sigma$ value changes the filter policy of the classifier, as it can prioritize samples closer to the evaluated sample when it's low and vice-versa. Figs~\ref{fig:sigma_comp_low} and~\ref{fig:sigma_comp_high} show how it can prioritize either distance or number of connections, and therefore change the value of $\text{q}_d(\textbf{X}_i)$. As it can be seen, as $\sigma \to \infty$, $\mathcal{K}(\textbf{X}_i, \textbf{X}_k) \to 1$ as $2\sigma^2 \gg ||\textbf{X}_i-\textbf{X}_k||^2$. Thus, $\text{q}_d(\textbf{X}_i)$ becomes the cardinality function of Eq.~\ref{eq:membership_value} since all kernel values will be equal to 1 and only neighborhood relations will be taken into account. Therefore, $\text{q}_d(\textbf{X}_i)$ can be seen as a generalization of $\text{q}(\textbf{X}_i)$, and may be optimized by tuning the hyperparameter $\sigma$ which determines the best filter policy for the dataset.

\begin{figure}[h]
\centering
\subfloat[]{\includegraphics[width=1.73in]{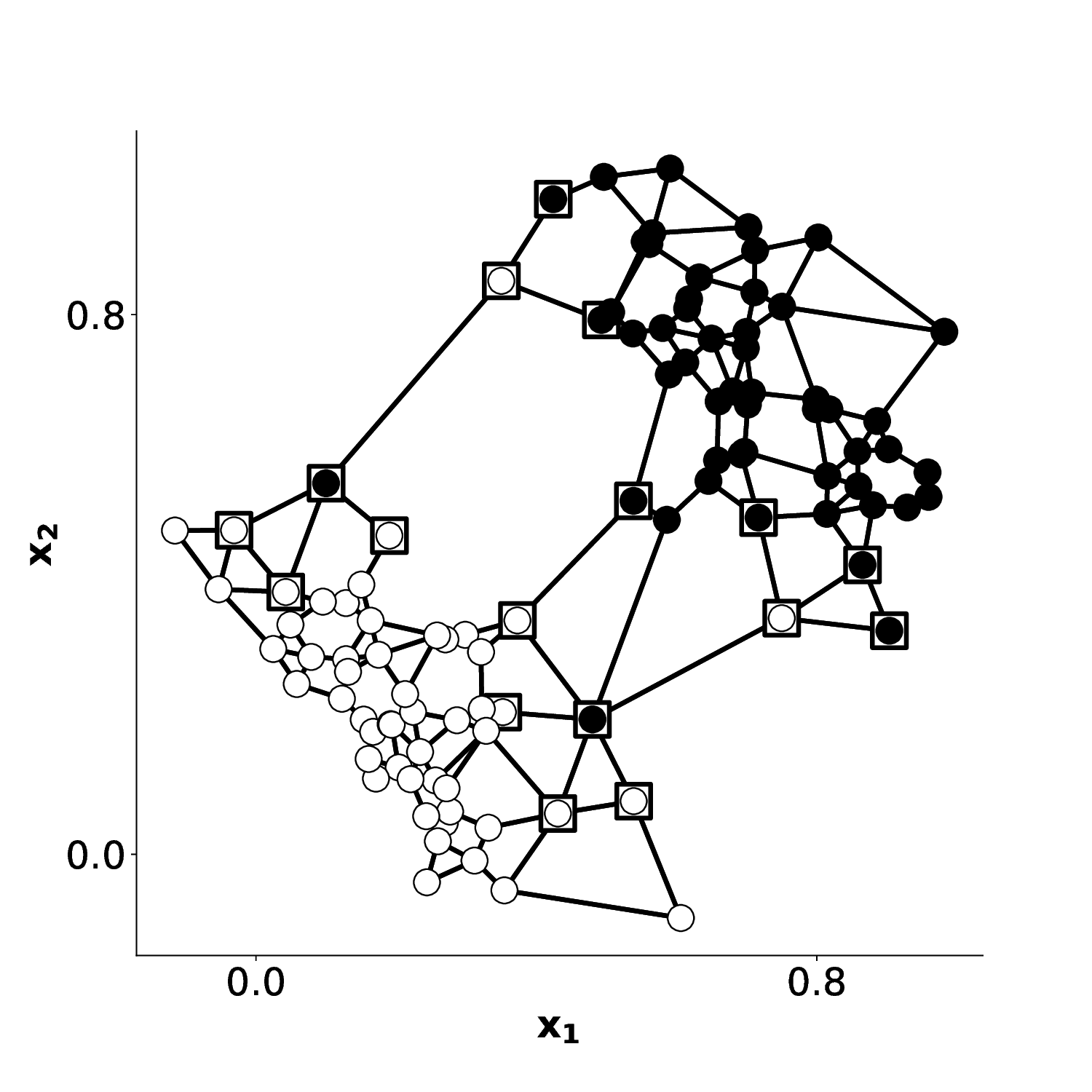}%
\label{fig:filter_out_chipclass}}
\hfil
\subfloat[]{\includegraphics[width=1.73in]{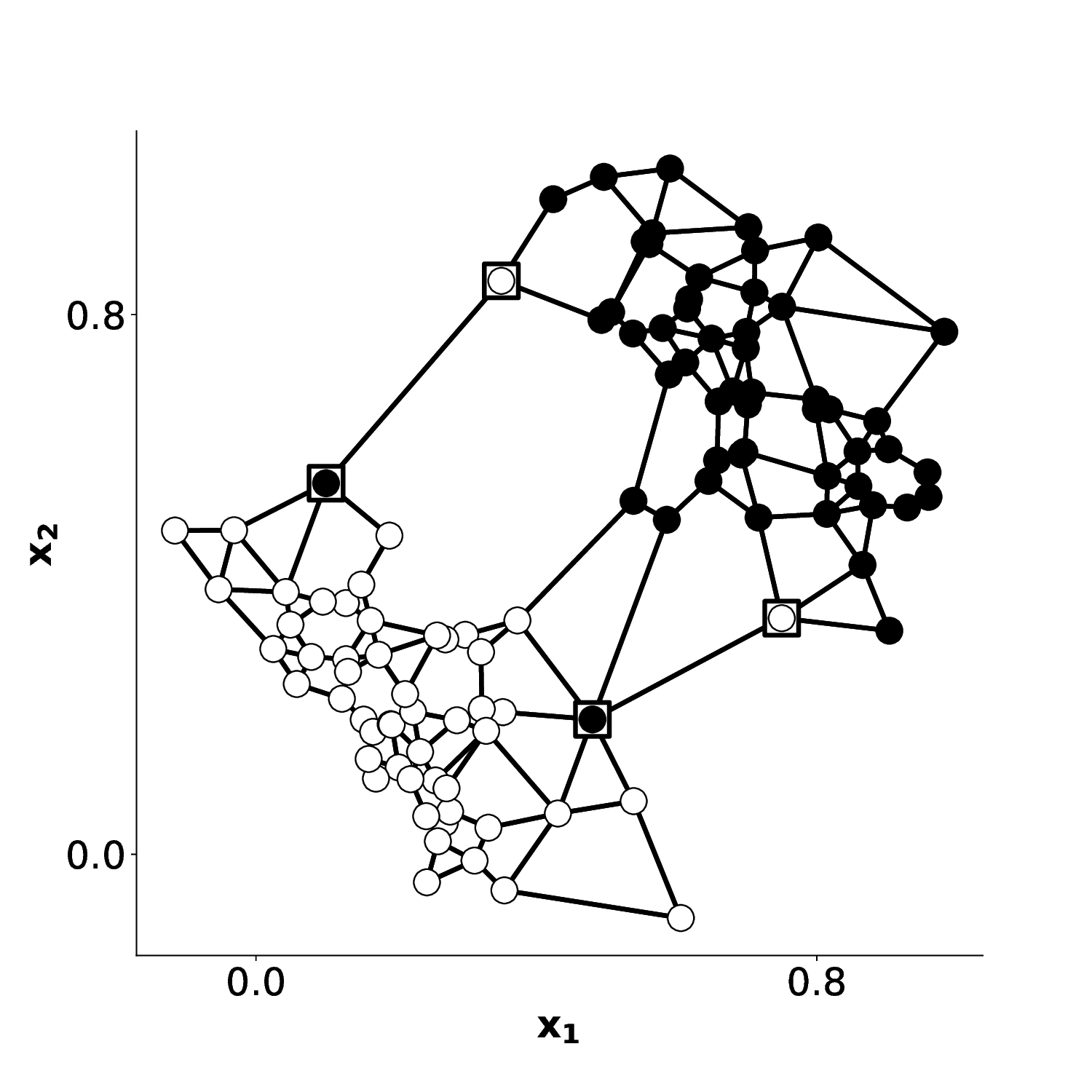}%
\label{fig:filter_out_distbased_sig}}
\caption{Filtered samples (in square) after applying class' thresholds defined in Eq.~\ref{eq:membership_threshold}, each figure using different quality values definitions. (a) $\text{q}(\textbf{X}_i)$ applied: outliers and their neighbors are filtered.  (b) $\text{q}_d(\textbf{X}_i)$ with $\sigma=0.03$ applied: only outliers are filtered.}
\end{figure}

Figs.~\ref{fig:filter_out_chipclass} and~\ref{fig:filter_out_distbased_sig} show how $\sigma$ can influence the performance of the classifier. While $\text{q}(\textbf{X}_i)$ penalizes samples neighboring the outliers, the new proposal allows $\sigma$ to be tuned so that only the outliers are filtered.

However, tuning $\sigma$ involves multiple GG-computations, as the GG must be computed after filtering and different $\sigma$s may filter different samples. The classic computation of the Gabriel Graph, following Eq.~\ref{eq:gg_formulation}, is depicted in algorithm~\ref{algo:gg_classic}. It traverses all possible distinct pairs of samples $j$ and $k$, which is equivalent to the upper triangular adjacency matrix $\ddot{G}$. For each pair, it traverses all samples to find if there's an $\textbf{X}_i$ that is within the $D$-sphere defined by $\textbf{X}_j$ and $\textbf{X}_k$, and if there is, it computes that $(j,k) \notin \ddot{G}$ and breaks the loop. 

\begin{algorithm}
\tiny
\caption{Classic computation of the Gabriel Graph}
\begin{algorithmic}
\STATE \textbf{Inputs}: $X$ \COMMENT{original dataset}
\STATE \textbf{Outputs}: $\ddot{G}$ \COMMENT{adjacency matrix}
\STATE $m \gets \textbf{length}(X)$
\STATE $\ddot{G} \gets$ ($m \times m$) matrix of 1s
\STATE $\ddot{G} \gets \ddot{G} - (m \times m)$ identity matrix
\FOR{$j=1$ to $m-1$}
    \FOR{$k=j+1$ to $m$}
        \STATE $d_{j,k} \gets ||X_j$-$X_k||^2$
        \FOR{$i=1$ to $m$}
            \IF{$d_{j,k} > ||X_j$-$X_i||^2 + ||X_k$-$X_i||^2$}
                \STATE $\ddot{G}_{j,k} \gets 0$
                \STATE $\ddot{G}_{k,j} \gets 0$
                \STATE \textbf{break}
            \ENDIF
        \ENDFOR
    \ENDFOR
\ENDFOR
\end{algorithmic}
\label{algo:gg_classic}
\end{algorithm}

Thus, the computational complexity of such algorithm can be considered as $\mathcal{O}(m^3)$, and recomputing the GG for $m-r$ samples, where $r$ is the number of filtered samples, would cost
\begin{equation}
    \mathcal{O}((m-r)^3)
    \label{eq:bigo_gg}
\end{equation}

\begin{figure}[H]
\centering
\subfloat[]{\includegraphics[width=1.73in]{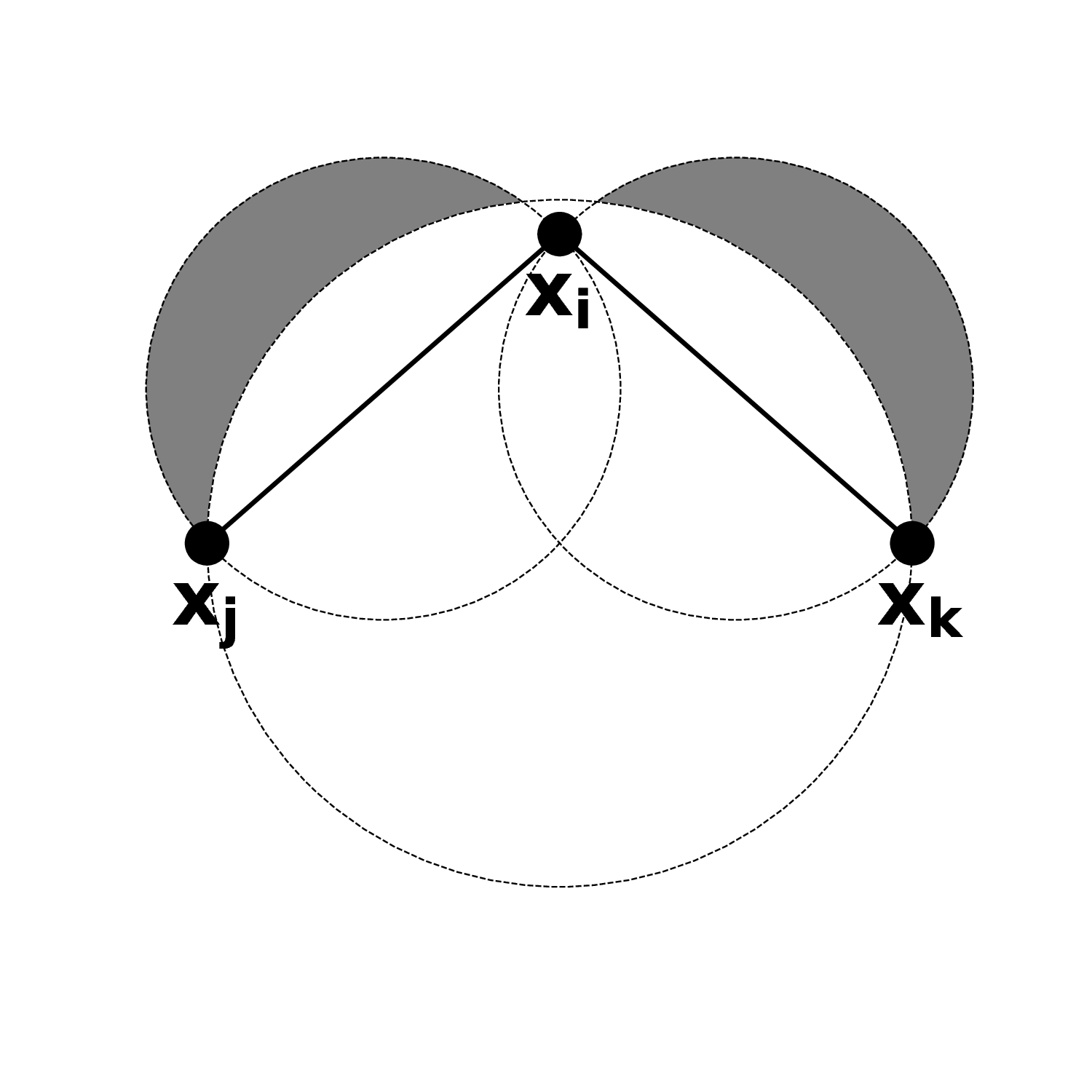}%
\label{fig:venn}}
\hfil
\subfloat[]{\includegraphics[width=1.73in]{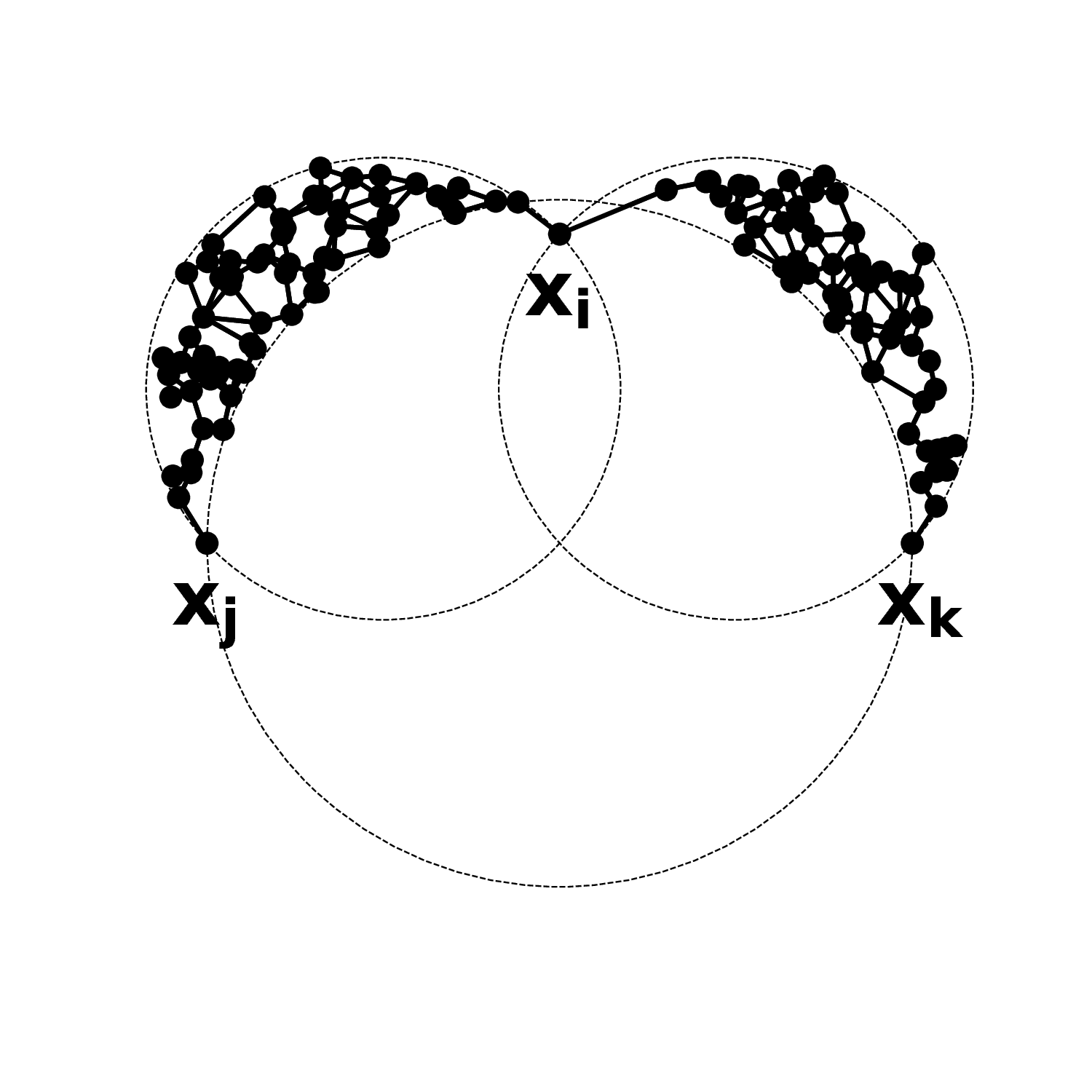}%
\label{fig:venngg}}
\caption{$\textbf{X}_i$ is within the $D$-sphere defined by $\textbf{X}_j$ and $\textbf{X}_j$, therefore ($\textbf{X}_j$,$\textbf{X}_k$) $\notin \mathcal{E}$ (a) Subsets $B-A$ and $C-A$ highlighted. (b) Example of a GG where multiple samples are between the samples ($\textbf{X}_k$,$\textbf{X}_i$) and ($\textbf{X}_i$,$\textbf{X}_j$).}
\end{figure}

However, such approach completely ignores the previous information obtained from the computation of the GG of the original set. One possibility to avoid such thing would be to recompute the graph based on the adjacency information between the nodes, not checking the effect of the removed sample to all the other samples but only the ones that are within some pre-defined node-to-node distance. However, consider two samples $\textbf{X}_j$ and $\textbf{X}_k$, and that there exists a sample $\textbf{X}_i$ that does not follow Eq.~\ref{eq:gg_formulation}, that is:
\begin{equation*}
    ||\textbf{X}_j-\textbf{X}_k||^2 > (||\textbf{X}_j-\textbf{X}_i||^2+||\textbf{X}_k-\textbf{X}_i||^2)
\end{equation*}
Let $A$ be the $D$-sphere centered at $(\textbf{X}_j+\textbf{X}_k)/2$ and diameter $||\textbf{X}_j-\textbf{X}_k||$ and $B$ the $D$-sphere centered at $(\textbf{X}_j+\textbf{X}_i)/2$ and diameter $||\textbf{X}_j-\textbf{X}_i||$. Considering the input space as $x \in \mathbb{R}^n$, if $B$ is not entirely within $A$, that is, 
\begin{equation*}
    \exists \ x \ (x \in B \land x \notin A)
\end{equation*}

\noindent which happens if
\begin{equation*}
    ||(\textbf{X}_j+\textbf{X}_k)/2-(\textbf{X}_j+\textbf{X}_i)/2|| + ||\textbf{X}_j-\textbf{X}_i||/2 > ||\textbf{X}_j-\textbf{X}_k||/2
\end{equation*}

\noindent then $B-A \neq \varnothing$, which means that there can exist $\infty$ samples within $B-A$ that do not affect the GG-edge between $\textbf{X}_j$ and $\textbf{X}_k$, however all these samples are between $\textbf{X}_j$ and $\textbf{X}_i$, requiring the traversal of an $\infty$-hop path to get from $\textbf{X}_j$ to $\textbf{X}_i$. As well as with $B$, the same holds for the $D$-sphere $C$ centered at $(\textbf{X}_k+\textbf{X}_i)/2$ and diameter $||\textbf{X}_k-\textbf{X}_i||$. Figs~\ref{fig:venn} and ~\ref{fig:venngg} illustrate such example, so that even though there can exist $\infty$ samples in the $B-A$ and $B-C$ subsets, if only $\textbf{X}_i$ is removed then ($\textbf{X}_j$,$\textbf{X}_k$) $\in \mathcal{E}$.

Thus, with the removal of one sample all non-adjacencies may still be checked. Another way to avoid to recompute the GG would be to store which samples were within the $D$-spheres defined by each pair of the dataset, and if all the samples were removed then that pair would become an edge of the GG. However, there are $m(m-1)/2$ pair possibilities and each possibility can have up to $m-2$ samples, which requires a large memory capacity, and besides that for each possibility a comparison between two arrays would also be required.

We propose a third approach. In the GG-computation of the original set, we store how many samples are within each $(j,k)$ pair in an upper triangular matrix $W$, as shown in algorithm~\ref{algo:gg_classic_in}. In the reconstruction for $m-r$ samples shown in algorithm~\ref{algo:sub_gg}, it's only needed to check how many samples from the $r$ subset are within the $D$-sphere of each $(j,k)$ pair. If this number of samples is equal to $W_{j,k}$, then all the samples within the $(j,k)$th $D$-sphere were removed, and then $(j,k) \in \ddot{G}$.  Thus, such approach would cost 
\begin{equation}
    \mathcal{O}(r(m-r)^2)
    \label{eq:bigonew_gg}
\end{equation}

\noindent in cases where $m \gg r$, which occurs when there are few outliers, then the complexity can be given as $\mathcal{O}(m^2)$.

\begin{algorithm}
\tiny
\caption{Computation of the Gabriel Graph while storing the number of samples within each (j,k) $D$-sphere}
\begin{algorithmic}
\STATE \textbf{Inputs}: $X$ \COMMENT{original dataset}
\STATE \textbf{Outputs}: $\ddot{G}$ \COMMENT{adjacency matrix}; $W$ \COMMENT{"within" matrix}
\STATE $m \gets \textbf{length}(X)$
\STATE $W \gets$ ($m \times m$) matrix of 0s
\STATE $\ddot{G} \gets$ ($m \times m$) matrix of 1s
\STATE $\ddot{G} \gets \ddot{G} - (m \times m)$ identity matrix
\FOR{$j=1$ to $m-1$}
    \FOR{$k=j+1$ to $m$}
        \STATE $d_{j,k} \gets ||X_j$-$X_k||^2$
        \FOR{$i=1$ to $m$}
            \IF{$d_{j,k} > ||X_j$-$X_i||^2 + ||X_k$-$X_i||^2$}
                \STATE $\ddot{G}_{j,k} \gets 0$
                \STATE $\ddot{G}_{k,j} \gets 0$
                \STATE $W_{j,k} \gets W_{j,k}+1$
            \ENDIF
        \ENDFOR
    \ENDFOR
\ENDFOR
\end{algorithmic}
\label{algo:gg_classic_in}
\end{algorithm}

\begin{algorithm}
\tiny
\caption{Computation of the sub-GG after filtering $r$ samples from the original dataset}
\begin{algorithmic}
\STATE \textbf{Inputs}: $\hat{X}$ \COMMENT{filtered samples}; $\tilde{X}$ \COMMENT{remaining samples}; $\tilde{W}$ \COMMENT{remaining samples' "within" matrix}
\STATE \textbf{Outputs}: $\tilde{\ddot{G}}$  \COMMENT{adjancecy matrix of the remaining samples}
\STATE $m_r \gets \textbf{length}(\tilde{X})$ \COMMENT{m-r}
\STATE $r \gets \textbf{length}(\hat{X})$
\STATE $\tilde{\ddot{G}} \gets$ ($m_r \times m_r$) matrix of 0s
\FOR{$j=1$ to $m_r-1$}
    \FOR{$k=j+1$ to $m_r$}
        \STATE $d_{j,k} \gets ||\tilde{X}_j$-$\tilde{X}_k||^2$
        \STATE $s \gets 0$
        \FOR{$i=1$ to $r$}
            \IF{$d_{j,k} > ||\tilde{X}_j$-$\hat{X}_i||^2 + ||\tilde{X}_k$-$\hat{X}_i||^2$}
                \STATE $s \gets s + 1$
            \ENDIF
            \IF{$s \textbf{=} \tilde{W}_{j,k}$}
                \STATE $\tilde{\ddot{G}}_{j,k} \gets 1$
                \STATE $\tilde{\ddot{G}}_{k,j} \gets 1$
            \ENDIF
        \ENDFOR
    \ENDFOR
\ENDFOR
\end{algorithmic}
\label{algo:sub_gg}
\end{algorithm}

\subsection{Architecture}
\label{subsec:architecture_svvoriented}

\subsubsection{Probability values near the margin}
After applying $\sum\limits_{k=1}^m\text{h}_k(\textbf{x})=1$, $0 < \text{h}_k(\textbf{x}) \leq 1$ and considering that $\text{max}(||\textbf{x}-\textbf{P}_i||) \gg ||\textbf{x}-\textbf{P}_k|| \; \forall \; i= 1,...,h$, the limit of $\text{h}_k(\textbf{x})$ with respect to $||\textbf{x}-\textbf{P}_k||$ is
\begin{equation*}
    \lim_{||\textbf{x}-\textbf{P}_k||\to0^+} \text{h}_k(\textbf{x}) = 1
\end{equation*}

\noindent that also applies to the activation function proposed in Eq~\ref{eq:gating_tanh}.

\begin{figure}[h]
\centering
\subfloat[]{\includegraphics[width=1.73in]{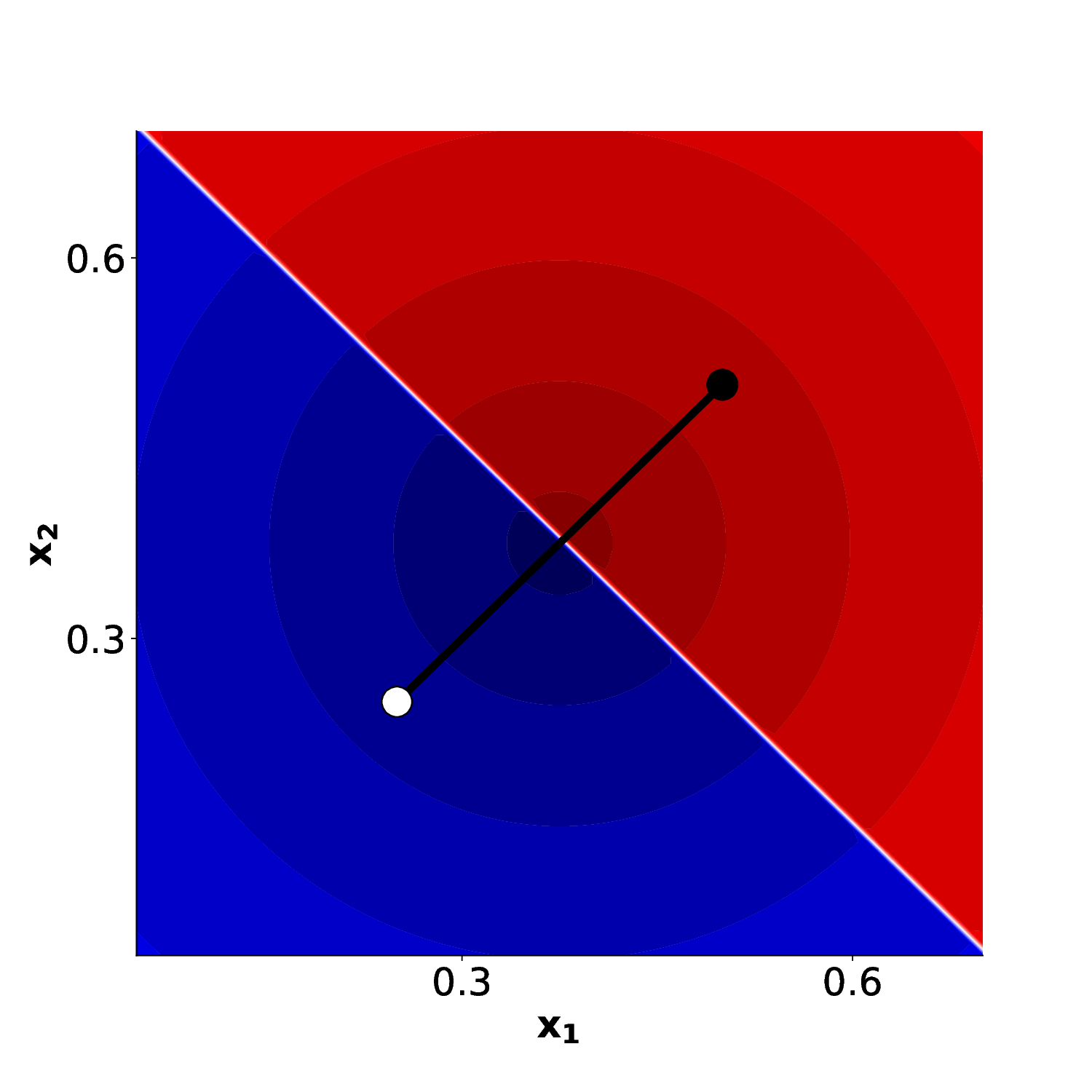}%
\label{fig:chipclass_margin}}
\hfil
\subfloat[]{\includegraphics[width=1.73in]{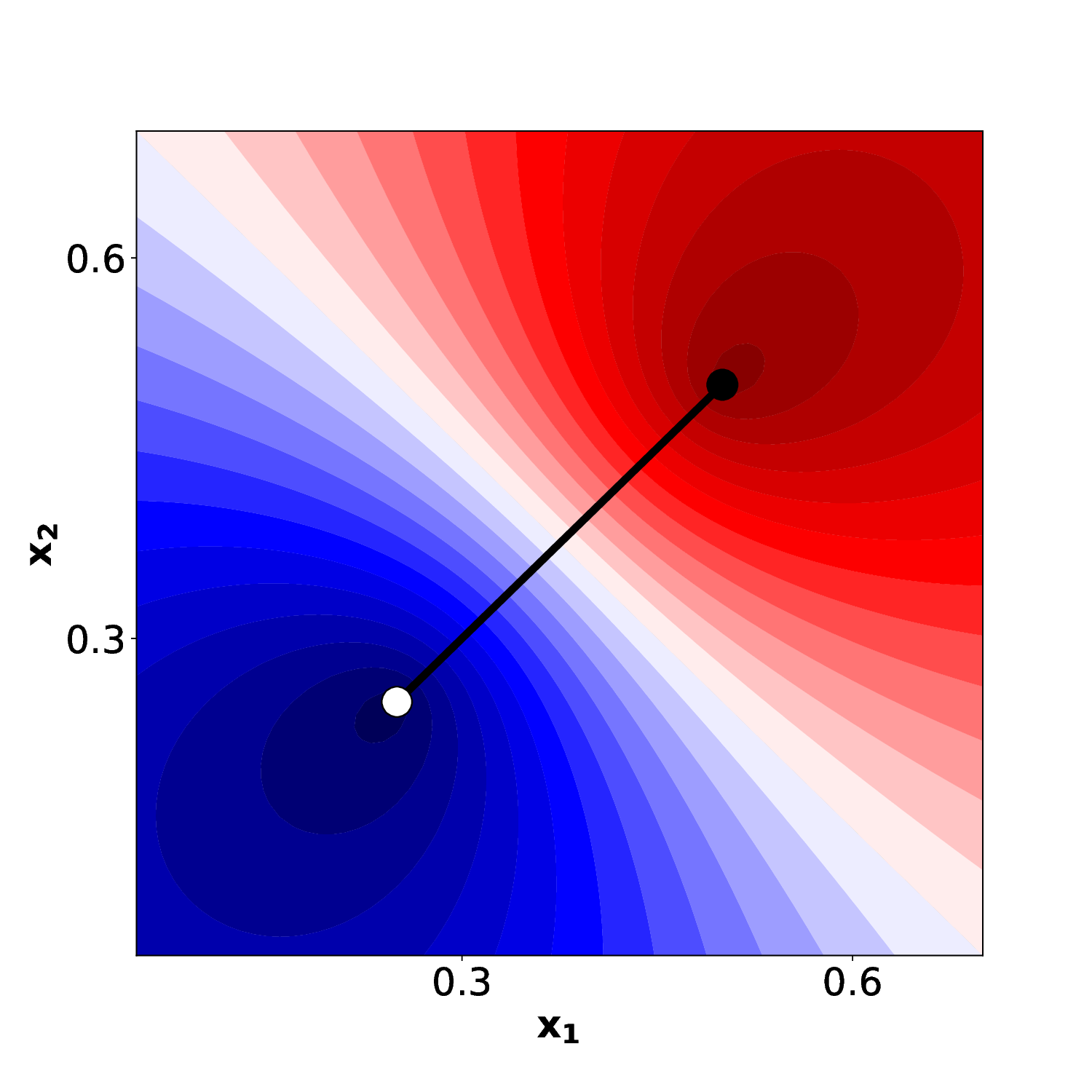}%
\label{fig:ssvchipclass_margin}}
\caption{(a) Chipclass' $\text{h}_{ktanh}(\textbf{x})$ density surface for 2 samples. (b) SSV-oriented Chipclass' $\text{h}_{ktanh}(\textbf{x})$ density surface for 2 samples.}
\end{figure}

\begin{figure}[h]
\centering
\subfloat[]{\includegraphics[width=1.73in]{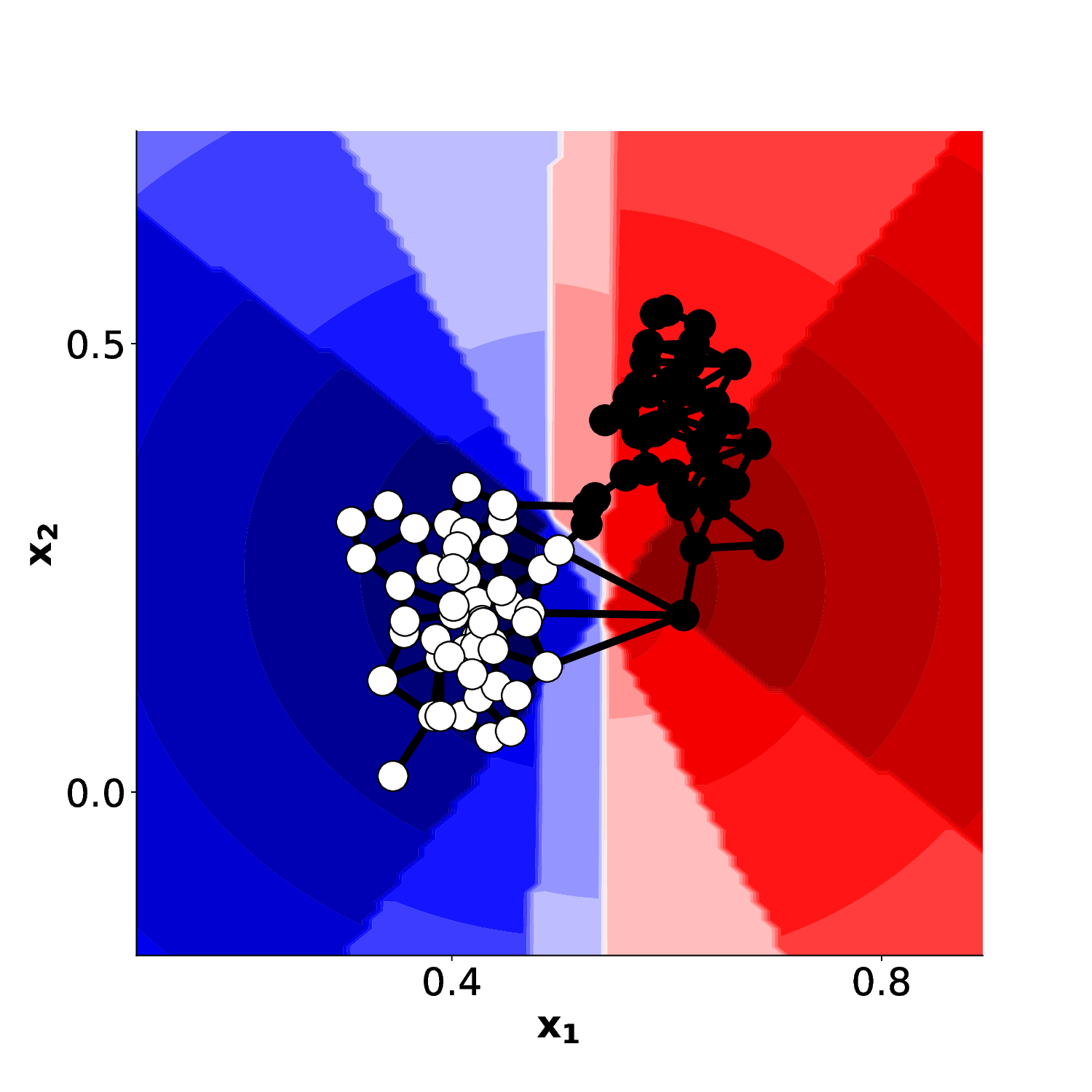}%
\label{fig:chipclass_discretization}}
\hfil
\subfloat[]{\includegraphics[width=1.73in]{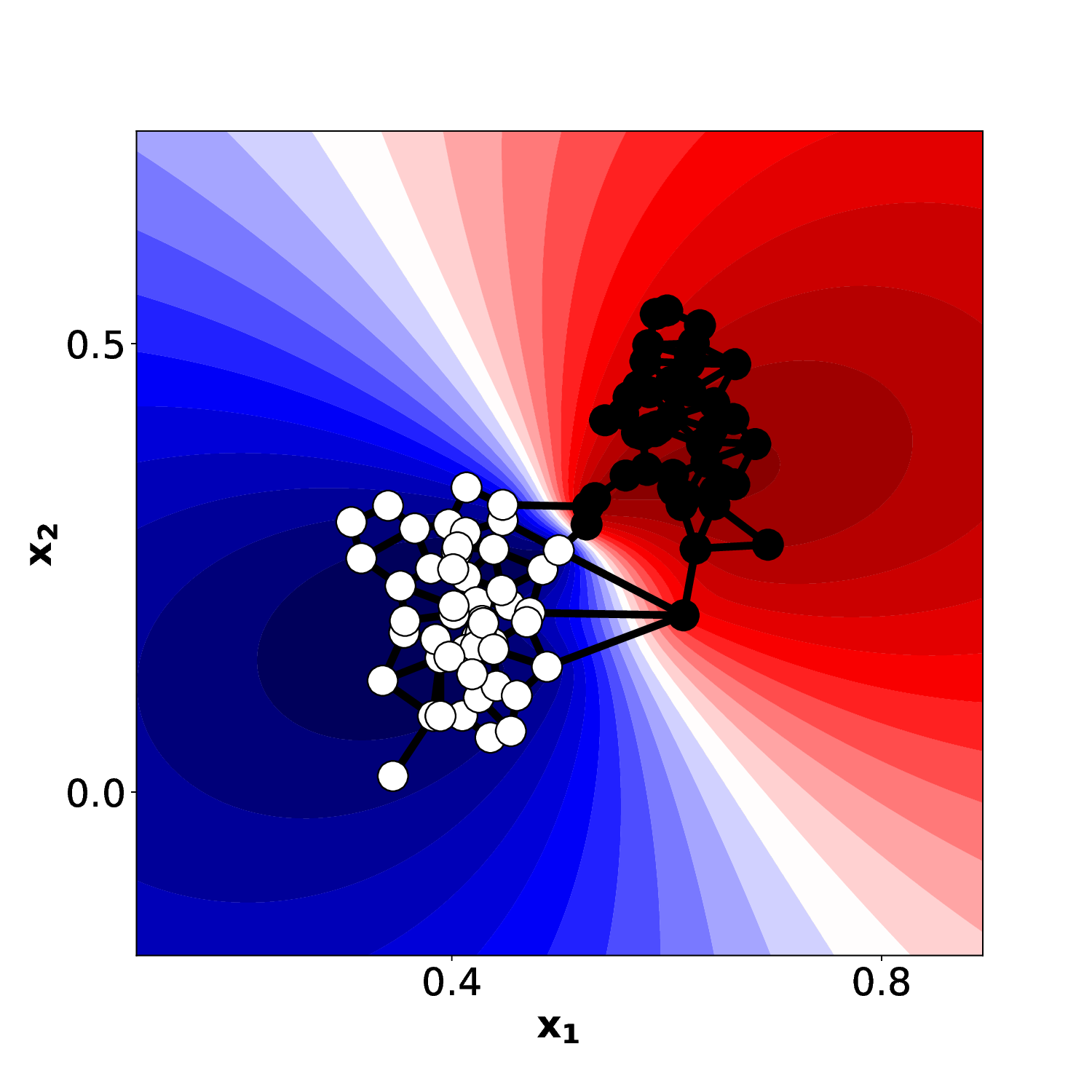}%
\label{fig:ssvchipclass_discretization}}
\caption{(a) Chipclass' $\text{h}_{ktanh}(\textbf{x})$ density surface for a binary classification problem (b) SSV-oriented Chipclass' $\text{h}_{ktanh}(\textbf{x})$ density surface for a binary classification problem.}
\end{figure}

Thus, the hyperplanes that are closer to $\textbf{x}$ have greater $\text{h}_k(\textbf{x})$ values and therefore contribute more to the final classification. However, since the activation functions are centered in $\textbf{P}_k$, the greatest probability values for a single hidden layer neuron are located in the margin between the two SSVs that define $\textbf{P}_k$, as shown in Fig.~\ref{fig:chipclass_margin}. Besides that, Chipclass can be seen as a linear combination of kNNs, where each hidden layer neuron corresponds to a kNN (with $k=1$), the neighbors used for the classifier are the $k$th pair of SSVs, as defined in Eq.~\ref{eq:definition_cpcn}, and the assignment of the weight of each kNN is given by the activation function, be it $\text{h}_k(\textbf{x})$ or $\text{h}_{ktanh}(\textbf{x})$. In this way, the discretization of $\textbf{w}_k$ leads to a discretization of the model's classification surface, as shown in Fig.~\ref{fig:chipclass_discretization}.

Nonetheless, one could center the activation functions on the SSVs instead of the midpoints, so that each neuron corresponds to an SSV. Thus, for a pair of SSVs, the highest densities are located close to the them, and on the margin the probabilities cancel out, as shown in Fig.~\ref{fig:ssvchipclass_margin}. Furthermore, the binarization imposed on Eq.~\ref{eq:definition_cpcn} can be discarded, which leads to smoother classification contours. Finally, as an SSV can be part of more than one SE, each SSV can be weighted according to Eq.~\ref{eq:pseudoinverse}.
\begin{equation}
    \textbf{w} = \textbf{H}^+ \textbf{Y} 
    \label{eq:pseudoinverse}
\end{equation}

\noindent where $\textbf{H}^+$ is the pseudo inverse of $\textbf{H}$. 

Thus, $\mathbb{P}(y=1|\textbf{x})$ remains as described in Eq.~\ref{eq:prob_chipclass}, however $\textbf{w}_k$ values are obtained from Eq.~\ref{eq:pseudoinverse} and $\text{h}_{ktanh}(\textbf{x})$ is presented in Eq.~\ref{eq:gating_tanh_ssvchipclass}.
\begin{equation}
    \text{h}_{ktanh}(\textbf{x}) = \text{tanh}(-||\textbf{x}-\zeta_k||)+1
    \label{eq:gating_tanh_ssvchipclass}
\end{equation}

\noindent where $s$ is the number of SSVs and $\zeta_k$ is the $k$th SSV of the training set.

Fig.~\ref{fig:ssvchipclass_discretization} depicts the density surface of a binary classification problem when the hidden layer uses SSVs to define the center of the activation functions. As it can be observed, by considering the SSV to locate hidden layer activation functions, smoother separation surfaces are obtained with lower likelihoods of both classes in the margin region.

\subsection{Multi-class classification}
\label{subsec:multiclass_classification}
For multi-class classification, the architecture presented in subsection~\ref{subsec:architecture_svvoriented} is extended by considering a linear output layer. $\textbf{W}_k$ can be trained with backpropagation, by using a softmax function as presented in Eq.~\ref{eq:softmax} with a cross-entropy loss, or by applying Eq.~\ref{eq:pseudoinverse}, where $\textbf{Y}$ is represented as a one-hot encoded version of $\textbf{y}$ of size $(m \times c)$.

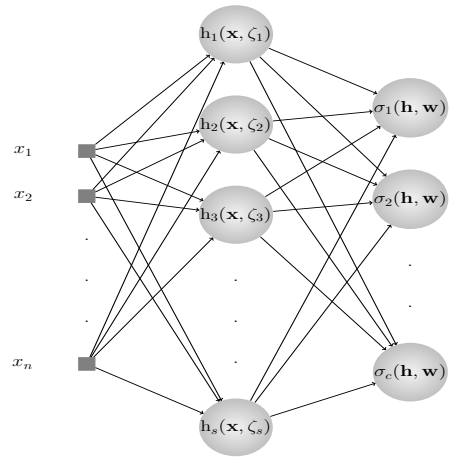
\begin{figure}[h]
	\centering
	\resizebox{6cm}{6cm}{
		\begin{tikzpicture}
			[rectin/.style={rectangle,draw=gray,fill=gray,thick,
				inner sep=0pt,minimum size=4mm},
			rectout/.style={rectangle,draw=none,fill=none,
				inner sep=0pt,minimum size=4mm},
			neuro/.style={circle,shade,inner color=gray!10,outer color=gray!50,thin,
				inner sep=0pt,minimum size=8mm},
			texto/.style={}]
			
			\node[rectin] (x1)       {};
			\node[rectin] (x2)  [below=of x1]     {};
			\node[] (x1txt)  [left=of x1]     {\Large $x_1$};
			\node[] (x2txt)  [left=of x2]     {\Large $x_2$};
			\node[] (pi1)  [below=of x2]     {\Large{.}};
			\node[] (pi2)  [below=of pi1]     {\Large{.}};
			\node[] (pi3)  [below=of pi2]     {\Large{.}};
			\node[rectin] (xn)  [below=of pi3]     {};
			\node[] (xntxt)  [left=of xn]     {\Large $x_n$};
			\node[neuro] (n1)  [above right=4cm of x1]     {\Large $\text{h}_1(\mathbf{x},\mathbf{\zeta}_1)$};
			\node[neuro] (n2)  [below= of n1]     {\Large $\text{h}_2(\mathbf{x},\mathbf{\zeta}_2)$};
			\node[neuro] (n3)  [below= of n2]     {\Large $\text{h}_3(\mathbf{x},\mathbf{\zeta}_3)$};
			
			\node[] (p1)  [below=of n3]     {\Large{.}};
			\node[] (p2)  [below=of p1]     {\Large{.}};
			\node[] (p3)  [below=of p2]     {\Large{.}};
			
			\node[neuro] (n5)  [below= of p3]     {\Large $\text{h}_s(\mathbf{x},\mathbf{\zeta}_s)$};

			\draw[->] (x1) edge (n1);
			\draw[->] (x1) edge (n2);
			\draw[->] (x1) edge (n3);
			\draw[->] (x1) edge (n5);
			\draw[->] (x2) edge (n5);
			\draw[->] (xn) edge (n5);

			\node[neuro] (nsaida)  [below right=1cm and 3cm of n1 ]     {\Large $\sigma_1(\mathbf{h},\mathbf{w})$};
			\node[neuro] (nsaida2)  [below=of nsaida]     {\Large $\sigma_2(\mathbf{h},\mathbf{w})$};
			\node[] (pif1)  [below=of nsaida2]     {\Large{.}};
			\node[] (pif2)  [below=of pif1]     {\Large{.}};
			\node[neuro] (nsaida3)  [below=of pif2]     {\Large $\sigma_c(\mathbf{h},\mathbf{w})$};
			
			\draw[->] (n1) edge (nsaida);
			\draw[->] (n2) edge (nsaida);
			\draw[->] (n3) edge (nsaida);
 			\draw[->] (n5) edge (nsaida);

			\draw[->] (n1) edge (nsaida2);
			\draw[->] (n2) edge (nsaida2);
			\draw[->] (n3) edge (nsaida2);
 			\draw[->] (n5) edge (nsaida2);
 
			\draw[->] (n1) edge (nsaida3);
			\draw[->] (n2) edge (nsaida3);
			\draw[->] (n3) edge (nsaida3);
 			\draw[->] (n5) edge (nsaida3);
 			
			\draw[->] (x2) -- (n1);
			\draw[->] (x2) -- (n2) ;
			\draw[->] (x2) -- (n3);

			\draw[->] (xn) -- (n1);
			\draw[->] (xn) -- (n2) ;
			\draw[->] (xn) -- (n3) ;

		\end{tikzpicture}

	}
	\caption{Schematic representation of SSV-oriented Chipclass for multi-class classification.}
	\label{fig:svvchipclass_multiclass}
\end{figure}

The architecture is depicted in Fig.~\ref{fig:svvchipclass_multiclass}.
\begin{equation}
    \sigma_k = \frac{\text{exp}(\textbf{W}_k^T\textbf{x})}{\sum\limits_{k=1}^c \text{exp}(\textbf{W}_k^T\textbf{x})}
    \label{eq:softmax}
\end{equation}

\begin{figure*}[t]
\centering
\includegraphics[scale=0.32]{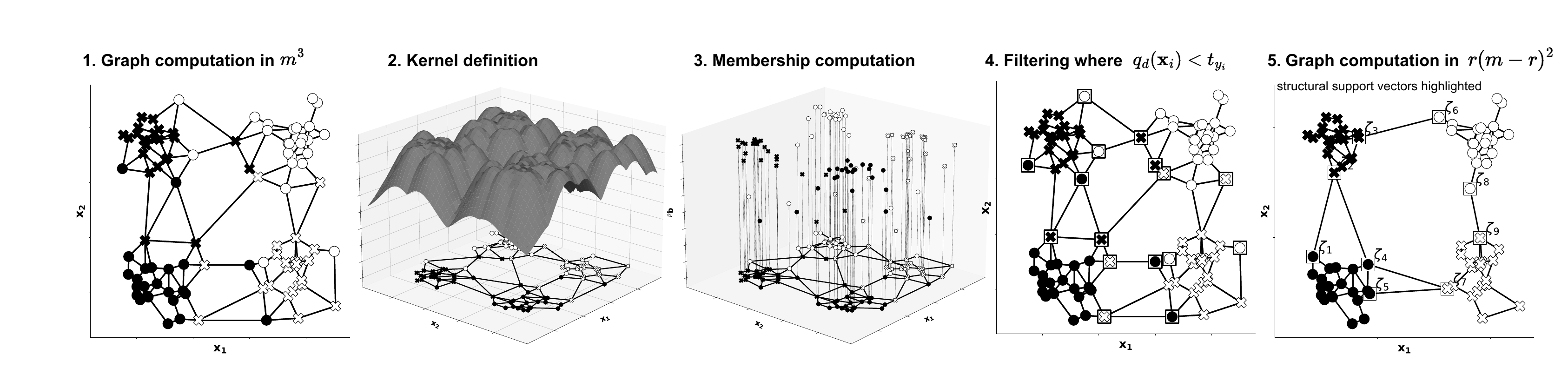}
\caption{
Step-by-step algorithm of the multi-class graph-based classifier proposed. (1) Gabriel graph computation in ($m^3$). (2) Definition of the kernels that follow Eq.~\ref{eq:gaussian_kernel}. (3) Computation of the membership function of Eq.~\ref{eq:membership_value_dist} based on the kernels previously defined. (4) Applying a filter based on each class' threshold and the membership function of each sample. (5) Recomputation of the GG in $r(m-r)^2$ and definition of the remaining SSVs.
}
\label{fig:schematic_multi}
\end{figure*}



Fig.~\ref{fig:schematic_multi} illustrates the walkthrough of the classifier proposed, including the kernel definition with the tuned $\sigma$ of Eq.~\ref{eq:gaussian_kernel} and consequently the membership function computation of Eq.~\ref{eq:membership_value_dist}, the recomputation of the GG after filtering the samples with the Algorithm~\ref{algo:sub_gg},
and the final SSVs used in the hidden layer of the the neural network that follows Fig.~\ref{fig:svvchipclass_multiclass} with the activation function proposed in Eq.~\ref{eq:gating_tanh_ssvchipclass}.

\section{Experiments and Results}
\label{sec:results}
\begin{figure}[h]
\centering
\includegraphics[width=3.5in]{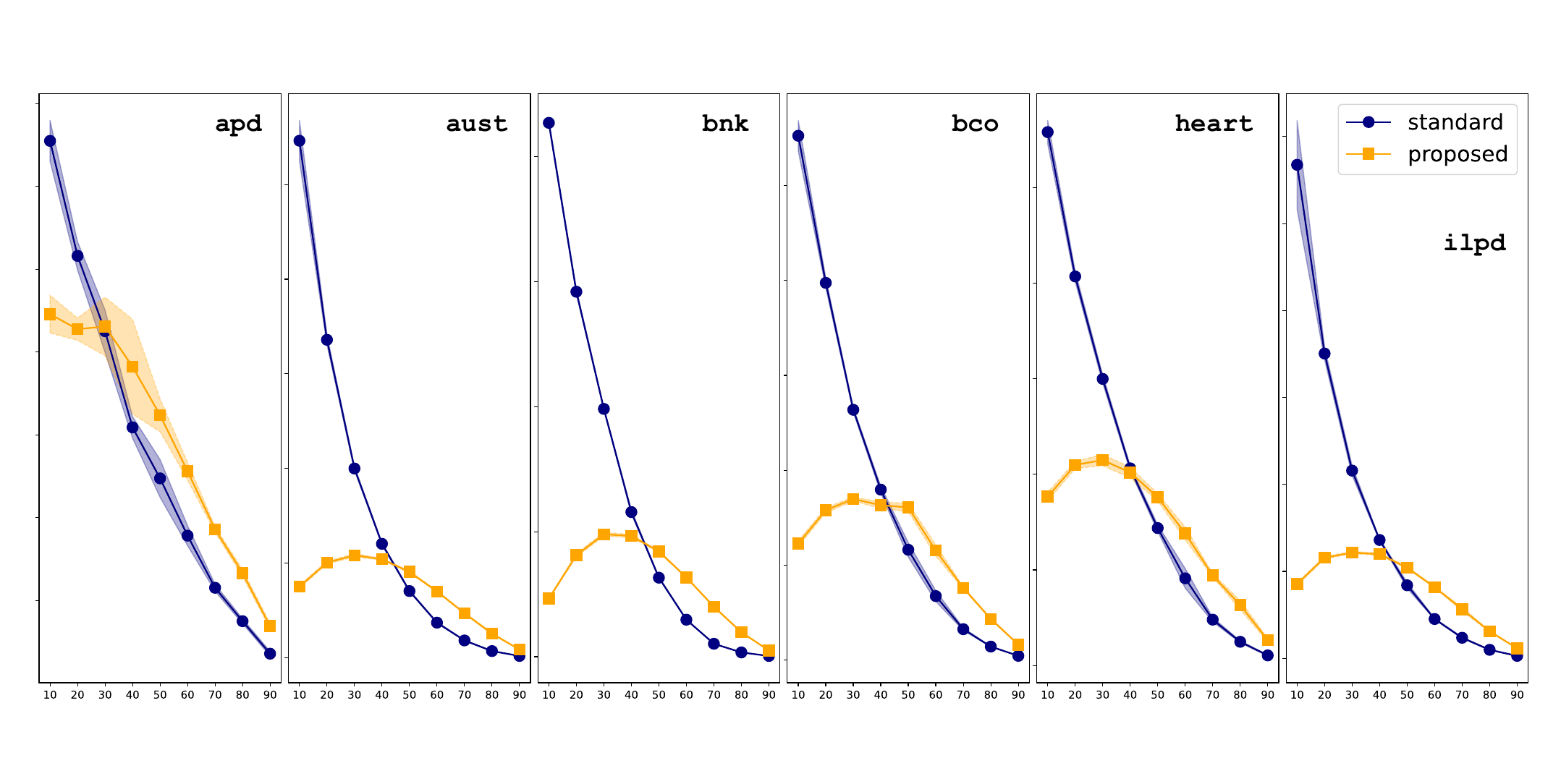}
\caption{Time taken to recompute GG after removing 10\% to 90\% of the samples of the datasets (x-axis). Each marker represents an average value of 10 runs at the percentage evaluated along with its uncertainty.}
\label{fig:study_recgg_time}
\end{figure}

Binary classification experiments were conducted with the Appendicitis dataset from the KEEL-dataset repository~\cite{alcala2011keel} and 15 datasets from the UCI repository~\cite{lichman2013uci}, 4 of them transformed into binary classification problems by either considering only two classes from the original set or applying one-versus-all, following the procedures adopted in~\cite{castro2013novel}. Details regarding number of attributes, number of samples and distributions of classes are presented in Table~\ref{tab:characteristics_datasets}. Multi-class classification experiments were conducted with 15 datasets previously used in the literature~\cite{hollmann2022tabpfn}.
\begin{table}[H]
\caption{Characteristics of the Datasets}
\label{tab:characteristics_datasets}
\centering
\scalebox{.8}{
\begin{tabular}{cccccc}
        \toprule
        \multicolumn{6}{c}{Binary classification datasets} \\ 
        \midrule
        \thead{Dataset} & Alias & \thead{No. of \\ attributes} & $m$ & $m:c_1$ & $m:c_2$ \\ 
        \midrule
        \thead{Appendicitis} & apd & 7 & 106 & 21 & 85 \\ 
        \thead{ILPD \\ (Indian Liver \\ Patient Dataset)} & ilpd  & 10 & 566 & 404 & 162 \\
        \thead{Australian Credit Approval} & aust & 14 & 690 & 307 & 383 \\ 
        \thead{Ionosphere} & iono  & 34 & 350 & 225 & 125 \\
        \thead{Banknote authentication} & bnk & 4 & 1348 & 610 & 738 \\ 
        \thead{Parkinsons} & par  & 22 & 195 & 147 & 48 \\
        \thead{Breast Cancer Wisconsin (Original)} & bco  & 9 & 449 & 236 & 213 \\
        \thead{Statlog (Vehicle Silhouettes)  4 vs. all} & v4  & 18 & 846 & 199 & 647 \\
        \thead{Breast Cancer Wisconsin  (Prognostic)} & bcp  & 32 & 194 & 148 & 46 \\ 
        \thead{Glass Identification  7 vs. all} & gls7  & 9 & 213 & 29 & 184 \\
        \thead{Climate Model  Simulation Crashes} & cli & 18 & 540 & 494 & 46 \\ 
        \thead{Yeast 5 vs. all}  & yst5  & 8 & 1453 & 51 & 1402 \\
        \thead{Fertility} & fer  & 9 & 98 & 87 & 11 \\ 
        \thead{Yeast 9 vs. 1} & yst9-1  & 8 & 458 & 20 & 438 \\ 
        \thead{Haberman's Survival} & hab  & 3 & 277 & 204 & 73 \\ 
        \thead{Abalone 18 vs. 9} & a18-9  & 10 & 731 & 42 & 689 \\
        \thead{Statlog (Heart)} & heart  & 13 & 270 & 150 & 120 \\ 
        \bottomrule
\end{tabular}
}
\end{table}
At first, Fig.~\ref{fig:study_recgg_time} depicts the time taken to compute a GG after removing $r$ samples for the standard and proposed approaches, described by algorithms~\ref{algo:gg_classic} and~\ref{algo:sub_gg} with computational complexities of Eqs.~\ref{eq:bigo_gg} and~\ref{eq:bigonew_gg}, respectively, for 6 datasets. As it can be seen, the proposed approach generally presents lower time to recompute GG before removing 50\% of the samples, which is supported by the fact that $r<m-r$.

\begin{figure}[h]
\centering
\includegraphics[width=3.5in]{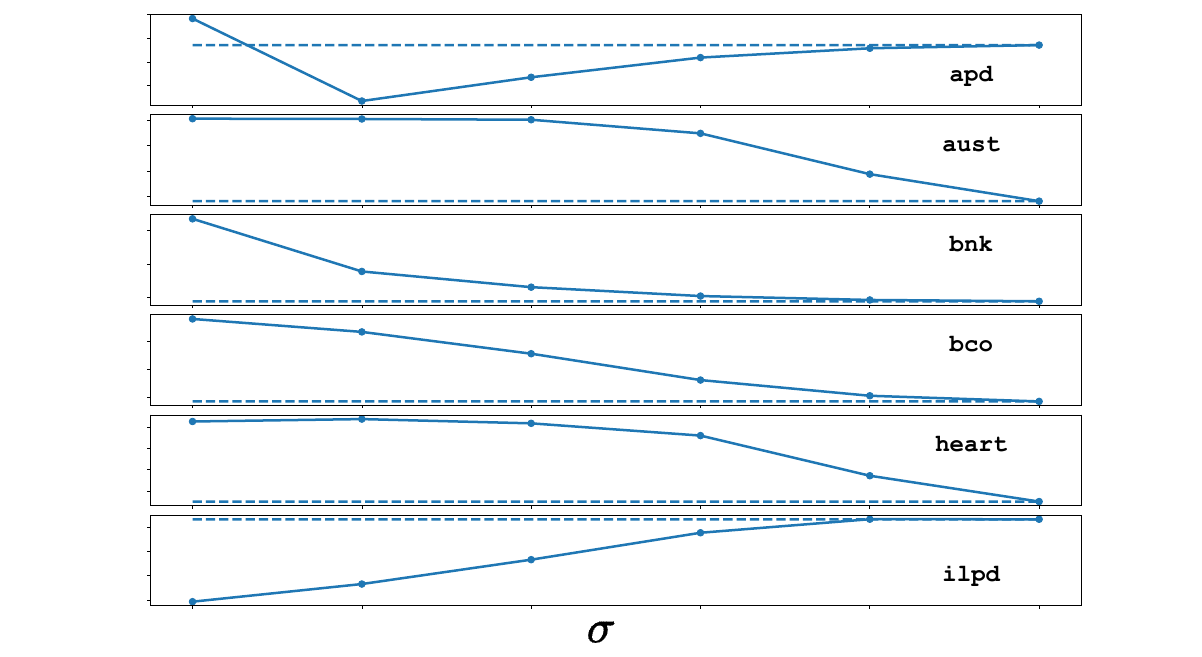}
\caption{Mean membership function of the entire dataset. Dashed line represents $\text{q}(\textbf{X}_i)$, whereas the solid line represents $\text{q}_d(\textbf{X}_i)$ varying with $\sigma$. As $\sigma \to \infty$, $\text{q}_d(\textbf{X}_i) \to \text{q}(\textbf{X}_i)$.}
\label{fig:sigmaeffect}
\end{figure}

Figure~\ref{fig:sigmaeffect} presents the $\sigma$ effect over the distance-based membership function proposed in Eq.~\ref{eq:membership_value_dist} for 6 datasets. As it can be seen, $\text{q}_d(\textbf{X}_i)$ is a generalization of $\text{q}(\textbf{X}_i)$ since it allows for different values of membership function depending on the kernel definition used in its computation. Meanwhile, if the kernel is too flat ($\sigma \to \infty$), then $\text{q}_d(\textbf{X}_i)$ becomes $\text{q}(\textbf{X}_i)$. Furthermore, Tab.~\ref{tab:chipclass_results} presents an ablation study for Chipclass when varying the activation function and the membership function used to compute the AUC of the test set. While $\text{q}(\textbf{X}_i)$ has a fixed membership function, Chipclass + $\text{q}_d(\textbf{X}_i)$ was computed with a grid search for $\sigma$, showing that the best result with $\text{q}_d(\textbf{X}_i)$ is always equal or better than the result using $\text{q}(\textbf{X}_i)$, as $\text{q}_d(\textbf{X}_i)$ is a generalization of $\text{q}(\textbf{X}_i)$ and a high $\sigma$ was included in the search. Chipclass with $\text{h}_{ktanh}$ also achieved a better avg. rank than $\text{h}_k$.

The following experiments were divided into 3 tables: Table~\ref{tab:ssvchipclass} presents SSV-oriented Chipclass with $\text{h}_{ktanh}(\textbf{x})$, compared to Chipclass, RBF-GG and GMM-GG. Tables~\ref{tab:binary_classifiers} and~\ref{tab:multiclassresults} present binary and multi-class classification results, respectively, for SSV-oriented Chipclass, kNN, SVMs, ResNets and tree-based methods: Random Forests, XGBoost and LightGBM. For all these experiments, nested cross-validation was applied: the dataset was split into 5 folds, each fold being evaluated with the model trained on the 80\% remaining samples. These 80\% samples were also used in a 5-fold stratified cross-validation for hyperparameter tuning, which was performed for all classifiers using Bayesian Optimization~\cite{snoek2012practical,akiba2019optuna}, following the hyperparameter space of ~\cite{shwartz2022tabular,hollmann2022tabpfn} for the literature models. kNN, SVMs and Random Forests were applied using Scikit-learn~\cite{pedregosa2011scikit}, while gradient boosting models followed their own libraries and ResNets were implemented based on the architecture proposed in~\cite{gorishniy2021revisiting}. For GG-based classifiers, $\sigma$ was tuned with a log-scaled uniform distribution ranging from 0.1 to 10. For Tables~\ref{tab:binary_classifiers} and~\ref{tab:multiclassresults}, SSV-oriented Chipclass had a per-class hyperparameter for the number of samples to filter from the first computation of the graph.

\begin{table}[h]
\caption{Ablation study: comparison of Chipclass models with cardinality ($\text{q}$) or distance-based ($\text{q}_d$) membership functions and standard ($\text{h}_k$) or smoother ($\text{h}_{ktanh}$) activation functions}
\label{tab:chipclass_results}
\centering
\scalebox{.8}{
\begin{tabular}{ccc|cc}
\toprule 
& \multicolumn{2}{c}{$\text{q}(\textbf{X}_i)$} & \multicolumn{2}{c}{$\text{q}_d(\textbf{X}_i)$ (best)} \\
\midrule
 & $\text{h}_k(\textbf{x})$ & $\text{h}_{ktanh}(\textbf{x})$ & $\text{h}_k(\textbf{x})$ & $\text{h}_{ktanh}(\textbf{x})$   \\
\midrule
a18-9 & 55.0571 & 73.847 & 61.9923 & 78.4356 \\
apd & 70.2778 & 81.3194 & 77.0833 & 83.6806 \\
aust & 85.9264 & 91.1915 & 91.7001 & 91.7362 \\
bnk & 99.7275 & 99.8089 & 99.7297 & 99.8222 \\
bco & 91.1665 & 93.2853 & 94.375 & 96.1592 \\
bcp & 57.8238 & 61.1762 & 65.8524 & 63.6095 \\
cli & 92.8867 & 94.0 & 94.9888 & 95.6612 \\
fer & 65.8333 & 65.8333 & 77.8472 & 71.5278 \\
gls7 & 95.2437 & 95.2827 & 97.8168 & 97.9825 \\
hab & 58.9341 & 69.3457 & 61.6849 & 72.4575 \\
heart & 83.2778 & 87.2222 & 84.5556 & 88.1111 \\
ilpd & 51.4056 & 67.3311 & 60.0813 & 69.4386 \\
iono & 95.0357 & 93.9024 & 95.2805 & 95.0264 \\
par & 73.4667 & 73.5762 & 92.019 & 93.6571 \\
v4 & 89.7319 & 94.7546 & 93.3926 & 96.2944 \\
yst5 & 62.0185 & 89.4807 & 65.3305 & 90.1511 \\
yst9-1 & 74.3235 & 74.8916 & 75.4598 & 75.4598 \\
\midrule
Avg. rank & 3.8529 & 2.6176 & 2.2647 & 1.2647 \\
\bottomrule
\end{tabular}
}
\end{table}

\begin{table}[h]
\caption{SSV-oriented Chipclass, RBF-GG and GMM-GG Comparison (Mean AUC 5 folds)}
\label{tab:ssvchipclass}
\centering
\scalebox{.8}{
\begin{tabular}{ccccc}
\toprule
 & Chipclass & RBF-GG & GMM-GG & SSV-oriented Chipclass \\
\midrule
a18-9 & 66.4994 & 88.6321 & 68.2888 & 88.305 \\
apd & 78.4118 & 79.1765 & 81.4706 & 85.9412 \\
aust & 91.2312 & 88.8418 & 91.7584 & 92.3255 \\
bnk & 99.8234 & 100.0 & 100.0 & 100.0 \\
bco & 96.0082 & 98.5613 & 98.899 & 97.3413 \\
bcp & 61.5479 & 54.1405 & 64.7867 & 56.7203 \\
cli & 93.527 & 90.5538 & 89.1418 & 91.4822 \\
fer & 60.2614 & 55.9477 & 73.7582 & 62.5817 \\
gls7 & 94.9499 & 95.6011 & 96.3248 & 96.7958 \\
hab & 71.3492 & 63.4702 & 63.9495 & 68.606 \\
heart & 88.0833 & 86.8056 & 90.0278 & 90.1667 \\
ilpd & 67.9805 & 63.0178 & 61.9819 & 66.9424 \\
iono & 94.4889 & 97.0667 & 92.96 & 98.2756 \\
par & 89.5326 & 96.9783 & 92.6718 & 94.1814 \\
v4 & 95.6632 & 99.7228 & 96.7266 & 99.1726 \\
yst5 & 90.0188 & 87.6656 & 84.4699 & 87.7037 \\
yst9-1 & 74.2117 & 74.1856 & 83.7565 & 83.0695 \\
\midrule
Avg. rank & 2.8824 & 2.8235 & 2.4706 & 1.8235 \\
\bottomrule
\end{tabular}
}
\end{table}
\begin{table}[h]
\caption{GG-based and Literature Models Comparison (Mean AUC 5 folds)}
\label{tab:binary_classifiers}
\centering
\scalebox{.6}{
\begin{tabular}{cccccccc}
\toprule
& kNN & SVM & LightGBM & Random Forest & XGBoost & ResNet & SSV-oriented Chipclass  \\
\midrule
a18-9 & 72.3743 & 92.0189 & 79.5674 & 80.5609 & 81.0135 & 92.9558 & 85.8953 \\
apd & 76.9706 & 82.9412 & 78.0588 & 79.0882 & 79.7941 & 74.0588 & 86.8235 \\
aust & 91.9287 & 91.9633 & 93.7844 & 93.9107 & 93.9663 & 93.1806 & 91.8667 \\
bnk & 99.8649 & 100.0 & 99.9978 & 99.9823 & 99.9978 & 100.0 & 100.0 \\
bco & 98.0014 & 98.9174 & 98.6621 & 98.3997 & 98.5827 & 98.9077 & 97.8702 \\
bcp & 59.1954 & 56.1175 & 58.5351 & 58.493 & 59.871 & 64.235 & 60.3167 \\
cli & 88.7541 & 95.535 & 94.635 & 92.5893 & 94.8289 & 89.9967 & 92.4025 \\
fer & 64.3301 & 52.7451 & 76.4052 & 73.2353 & 68.9869 & 55.5882 & 66.7647 \\
gls7 & 95.0841 & 95.8869 & 95.1929 & 96.1216 & 97.6917 & 92.4735 & 97.8769 \\
hab & 63.9567 & 73.3879 & 70.2361 & 67.9485 & 69.9918 & 67.1425 & 68.6236 \\
heart & 89.0139 & 91.2778 & 91.6389 & 91.4722 & 91.5556 & 88.5556 & 90.0 \\
ilpd & 64.4786 & 66.4408 & 72.2462 & 72.7652 & 73.1701 & 73.456 & 66.1625 \\
iono & 92.9689 & 97.9733 & 98.3822 & 97.5378 & 98.0622 & 97.3156 & 97.7956 \\
par & 96.9093 & 96.3295 & 97.3487 & 95.7752 & 96.1456 & 95.2286 & 96.3295 \\
v4 & 96.8944 & 99.8194 & 99.6094 & 99.4137 & 99.7625 & 99.6105 & 99.4177 \\
yst5 & 87.1444 & 84.9048 & 92.2925 & 92.4483 & 93.0076 & 85.2494 & 86.5513 \\
yst9-1 & 80.1832 & 80.7171 & 65.8177 & 89.4475 & 84.5461 & 77.0768 & 84.7753 \\
\midrule
Avg. rank & 5.8235 & 3.5 & 3.4412 & 4.0 & 2.7353 & 4.5882 & 3.9118 \\
\bottomrule
\end{tabular}
}
\end{table}

Significance tests were made using the Friedman test~\cite{friedman1937use}, a non-parametric ranking-based test that is the most suitable for the comparison of multiple classifiers over multiple datasets according to~\cite{demvsar2006statistical}. $F(L-1,(L-1)(N-1))$ were extracted from the F-distribution, which can be consulted in~\cite{sheskin2003handbook}.

For Tabs.~\ref{tab:binary_classifiers} and~\ref{tab:multiclassresults}, $F_F=4.2172>F(6,96)=2.1945$ and $F_F=3.2903>F(6,84)=2.2086$ ($\alpha=0.05$), respectively, and therefore the Bonferroni-Dunn test was applied. SSV-oriented Chipclass was within the range of the critical value and therefore statistically equivalent to the models present in the literature. For Tab.~\ref{tab:ssvchipclass} $F_F<F$ and therefore the null hypothesis that the classifiers are statistically equivalent can not be rejected. However, it can be seen that SSV-oriented Chipclass presented lower average ranks than Chipclass, RBF-GG and GMM-GG.

\begin{table}[H]
\caption{Mean ROC-AUC OvO of the 5 folds for the Multi-Class Classifiers}
\label{tab:multiclassresults}
\centering
\scalebox{.7}{
\begin{tabular}{cccccccc}
\toprule
OpenML ID & kNN & SVM & LightGBM & Random Forest & XGBoost & ResNet & \thead{SSV-oriented \\ Chipclass} \\
\midrule
1100 & 57.3892 & 59.6778 & 65.774 & 65.5689 & 65.3498 & 62.5485 & 62.5559 \\
1499 & 99.3027 & 99.5068 & 98.9796 & 98.733 & 98.8605 & 99.6259 & 99.3537 \\
1512 & 59.6342 & 59.1149 & 60.2289 & 62.054 & 62.3863 & 60.5407 & 62.6216 \\
1523 & 88.6356 & 94.5889 & 92.3093 & 92.9565 & 92.7167 & 93.6991 & 91.7889 \\
187 & 99.8233 & 99.9762 & 100.0 & 99.9798 & 99.96 & 99.8333 & 99.9798 \\
329 & 76.9861 & 97.8333 & 76.2143 & 96.1667 & 99.5476 & 85.0516 & 81.9266 \\
40682 & 98.6058 & 99.963 & 99.8386 & 99.8228 & 99.8704 & 99.8254 & 99.8122 \\
41 & 87.1792 & 91.1694 & 93.4614 & 95.3011 & 94.3082 & 89.7734 & 92.0703 \\
41919 & 70.6178 & 72.7291 & 70.4623 & 71.9767 & 72.0569 & 71.65 & 72.087 \\
42261 & 99.2778 & 99.8333 & 98.9667 & 99.4333 & 99.3593 & 100.0 & 99.9 \\
42544 & 94.8831 & 98.0725 & 97.2697 & 97.6477 & 97.4483 & 97.1691 & 97.5226 \\
48 & 62.8288 & 58.7642 & 50.1361 & 60.5896 & 63.4694 & 63.5941 & 59.9546 \\
61 & 99.7296 & 99.963 & 98.3037 & 99.4667 & 98.9296 & 100.0 & 99.8963 \\
679 & 72.8085 & 72.7442 & 67.8501 & 74.7854 & 59.0882 & 68.0544 & 73.1631 \\
694 & 99.9436 & 100.0 & 99.9681 & 99.99 & 99.939 & 99.8351 & 99.9901 \\
\midrule
Avg. rank & 5.6667 & 3.0667 & 4.8667 & 3.3 & 3.8667 & 3.8667 & 3.3667 \\
\bottomrule
\end{tabular}
}
\end{table}


\section{Conclusions}
\label{sec:conclusion}
The use of geometric structures of the dataset to obtain support vectors is an alternative to the classical approach of obtaining these points near the margin with quadratic programming, as in SVMs. Thus, such architectures were explored as well as the effects of the original strategies on the resulting separation surface.

It has been proposed the use of smooth activation functions for Chipclass, giving more relevance to the distance of the test sample to the center of the hidden layer neuron and avoiding a frequent disregard of hidden layer neurons. It was presented a new membership function that takes into account not only the configuration of the graph, but also the distances of the sample being evaluated to its neighbors. As it is based on a predefined $\sigma$ that may be tuned to achieve the best performance, a new algorithm to recompute the GG after filtering $r$ samples in $\mathcal{O}(r(m-r)^2)$ instead of $\mathcal{O}((m-r)^3)$ was proposed. It was discussed how activation functions centered on structural support vectors (SSVs) instead of boundary hyperplanes leads to a margin with low probabilities and smoother classification contours, with the new SSV-oriented Chipclass. Considering such principles, an extended neural network architecture for multi-class classification with a linear layer in the output layer was presented, with weights computed with gradient-descent and backpropagation or with the pseudo-inverse algorithm from RBF Networks. 

Statistical analysis with the Friedman test showed that Chipclass with smoother activation functions obtained better results than standard Chipclass. Moreover, SSV-oriented Chipclass was statistically equivalent to literature models and presented lower average ranks than other GG-based models.

\bibliographystyle{IEEEtran}
\bibliography{ref.bib}

\begin{IEEEbiographynophoto}{Vítor M. Hanriot}
received the B.Sc. degree in Control and Automation Engineering (2021) and the M.Sc. degree in Electrical Engineering (2023) from the Universidade Federal de Minas Gerais (UFMG).
\end{IEEEbiographynophoto}
\begin{IEEEbiographynophoto}{Luiz C. B. Torres}
received the Ph.D. (2016) and master’s (2012) degree in Electrical Engineering, both from UFMG, Brazil. His B.Sc degree in Computer Science was obtained from the University Center of Belo Horizonte (2010). Since 2019, he has been with the Computing and Systems Department at UFOP, Brazil.
\end{IEEEbiographynophoto}
\begin{IEEEbiographynophoto}{Antônio P. Braga}
received his Ph.D. in 1995, from Imperial College, London, in recurrent neural networks. Presently, he is a Professor at UFMG, where he heads the Computational Intelligence Lab. He's published many papers and books, supervised around 100 post-graduated students and served in editorial boards of international journals.
\end{IEEEbiographynophoto}

\vfill

\end{document}